\documentclass[lettersize,journal]{IEEEtran}
\usepackage{amsmath,amsfonts}
\usepackage{algorithmic}
\usepackage{array}
\usepackage[caption=false,font=normalsize,labelfont=sf,textfont=sf]{subfig}
\usepackage{indentfirst}
\usepackage{authblk}
\usepackage{notoccite}
\usepackage{textcomp}
\usepackage{stfloats}
\usepackage{url}
\usepackage{hyperref}
\usepackage{verbatim}
\usepackage{ulem}
\usepackage{bm}
\usepackage{graphicx}
\def\BibTeX{{\rm B\kern-.05em{\sc i\kern-.025em b}\kern-.08em
    T\kern-.1667em\lower.7ex\hbox{E}\kern-.125emX}}
\usepackage{mathtools}
\DeclareSymbolFont{matha}{OML}{txmi}{m}{it}
\DeclareMathSymbol{\varv}{\mathord}{matha}{118}
\usepackage{soul} 
\usepackage{amsmath,array,float}
\usepackage{balance}

\usepackage{booktabs,       
            makecell,       
            tabularx}       
\usepackage{multirow}
\usepackage{multicol}
\usepackage[export]{adjustbox}
\usepackage{balance}

\usepackage{xcolor}

\newcommand{\gmkn}{g_{mk}^{(n)}}
\newcommand{\hmkn}{h_{mk}^{(n)}}

\usepackage{tikz}

\usetikzlibrary{shapes.geometric}
\usetikzlibrary{arrows}
\usetikzlibrary{fit}
\usetikzlibrary{calc}
\tikzstyle{startstop} = [rectangle, rounded corners, minimum width=3cm, minimum height=1cm,text centered, text width=3cm, draw=black]
\tikzstyle{io} = [trapezium, trapezium left angle=70, trapezium right angle=110, minimum width=3cm, minimum height=1cm, text centered, text width=3cm, draw=black]
\tikzstyle{process} = [rectangle, minimum width=3cm, minimum height=1cm, text centered, text width=3cm, draw=black]
\tikzstyle{decision} = [diamond, aspect=3, align=center, inner sep=-1ex, minimum width=2cm, minimum height=1cm, text centered, text width=5cm, draw=black]
\tikzstyle{arrow} = [thick,->,>=stealth]

\def\BibTeX{{\rm B\kern-.05em{\sc i\kern-.025em b}\kern-.08em
    T\kern-.1667em\lower.7ex\hbox{E}\kern-.125emX}}
\begin{document}
\title{Deep Learning Based Approach for User Activity Detection with Grant-Free Random Access in Cell-Free Massive MIMO}
\author{
    Ali~Elkeshawy,~\IEEEmembership{Student Member,~IEEE},
    Ha\"{i}fa~Far\`{e}s,~\IEEEmembership{Member,~IEEE},
    and~Amor~Nafkha,~\IEEEmembership{Senior Member,~IEEE}
    \thanks{
        All authors are with the IETR UMR CNRS 6164, CentraleSupélec, Cesson Sévigné 35576, France. 
        Emails: \{ali-fekry-ali-hassan.elkeshawy, haifa.fares, amor.nafkha\}@centralesupelec.fr.
    }
    \thanks{This work received funding from the French National Research Agency (ANR) under grant ANR-22-CE25-0015 within the frame of the project POSEIDON.}
}


\maketitle

\begin{abstract}
Modern wireless networks must reliably support a wide array of connectivity demands, encompassing various user needs across diverse scenarios. Machine-Type Communication (mMTC) is pivotal in these networks, particularly given the challenges posed by massive connectivity and sporadic device activation patterns. Traditional grant-based random access (GB-RA) protocols face limitations due to constrained orthogonal preamble resources. In response, the adoption of grant-free random access (GF-RA) protocols offers a promising solution.
This paper explores the application of supervised machine learning models to tackle activity detection issues in scenarios where non-orthogonal preamble design is considered. We introduce a data-driven algorithm specifically designed for user activity detection in Cell-Free Massive Multiple-Input Multiple-Output (CF-mMIMO) networks operating under GF-RA protocols. Additionally, this study presents a novel clustering strategy that simplifies and enhances activity detection accuracy, assesses the resilience of the algorithm to input perturbations, and investigates the effects of adopting floating-to-fixed-point conversion on algorithm performance.
Simulations conducted adhere to 3GPP standards, ensuring accurate channel modeling, and employ a deep learning approach to boost the detection capabilities of mMTC GF-RA devices. The results are compelling: the algorithm achieves an exceptional 99\% accuracy rate, confirming its efficacy in real-world applications.

\end{abstract}


\section{Introduction}
\label{sec:introduction}
\IEEEPARstart{A}{n} important step toward adopting Internet of Things (IoT) applications has been taken with the introduction of mMTC services \cite{Bockelmann2016}. This marks the beginning of a new era where devices, constrained by the need for energy efficiency and limited battery life, transmit small amounts of data either sporadically  or regularly at short intervals. As these devices proliferate and become distributed across large regions, they present significant challenges to current network connectivity models related to the densification of data traffic.

Within this framework, massive MIMO networks are part of the solution that can overcome mMTC challenges by improving network coverage and supporting a higher device density \cite{Bana2019},\cite{Marzetta2010}.

Investigations have been conducted on traditional grant-based massive random access techniques, in which active devices choose an orthogonal pilot sequence to notify the base station of their desire to transmit data \cite{Pratas2012, Sorensen2014, Bjornson2016}. However, mutually orthogonal preamble sequences are hard to come by in wireless systems due to resource shortages caused by the short channel coherence time compared to the high number of devices to be served. On another side, one characteristic of mMTC is its sporadic traffic patterns,  combined with a significant collision risk among devices which might be actively using the network and small payload sizes. This situation frequently results in a large overlap in the sequences that different devices choose to use, which causes resource conflict and network collisions and expands access delays since re-transmissions are required. Furthermore, a significant amount of signaling overhead is involved in the collision resolution process, which disproportionately outweighs the short data payloads that devices broadcast in the context of mMTC applications~\cite{Ding2020}.

\indent Many grant-free access strategies have been presented to solve the limitations of grant-based random access mechanisms\cite{Shahab2020}. These solutions allow devices to join a wireless network without express permission. In these grant-free applications, every device is assigned a unique non-orthogonal pilot sequence, avoiding the challenges of orthogonal sequence assignment because of the short channel coherence time. In light of this, grant-free random access is seen as a feasible method to guarantee network connectivity by effectively controlling non-orthogonal preamble interference and reducing network access latency \cite{Chen2021}.
 For effective data communication in mMTC environments, precise channel estimation is essential, necessitating the identification of active users, a process significantly influenced by the sparse activity emitted by machine-type devices. Our investigation focuses on sparse support recovery techniques designed to identify the active device indices within a transmission slot under a grant-free access model.

\begin{figure}[h]
  \centering
  \includegraphics[width=\columnwidth]{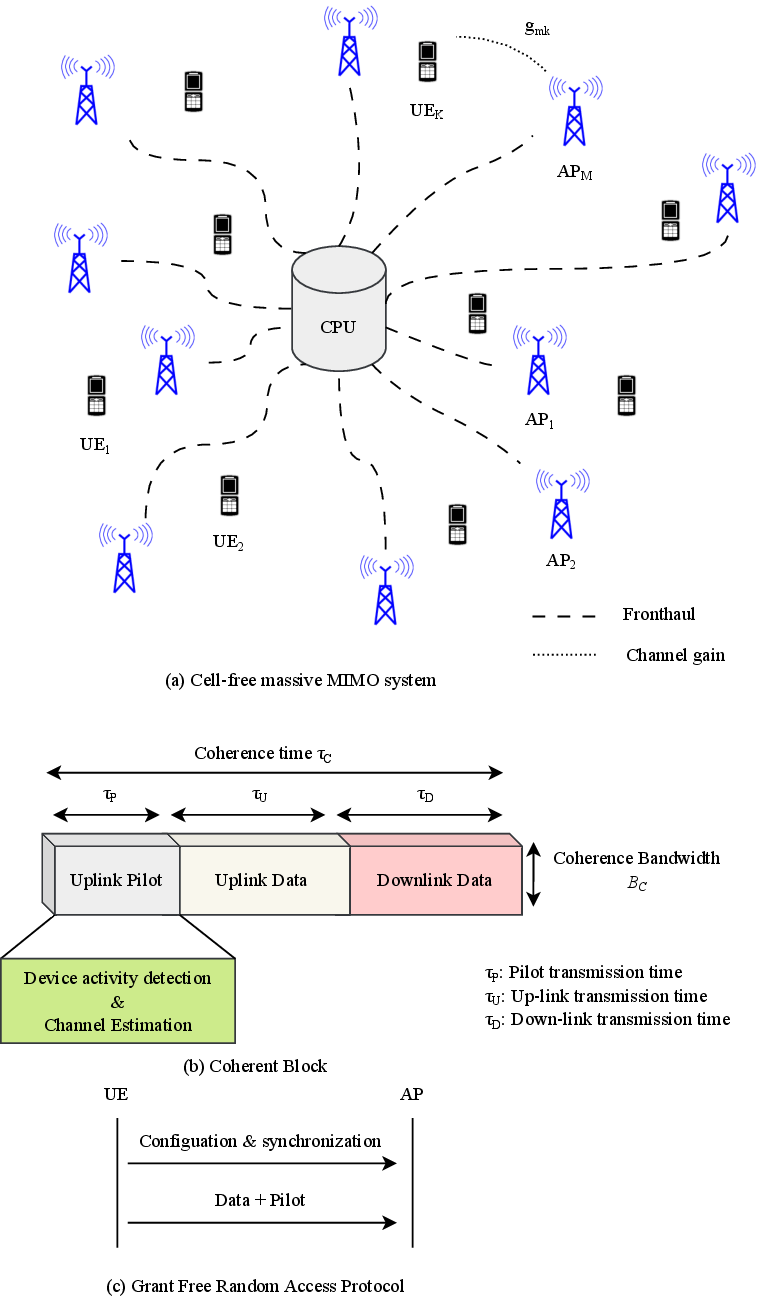}
  \caption{Cell-Free network model for mMTC.}
  \label{fig1}
\end{figure}

\indent The methodologies to perform sparse support recovery can be classified into three main categories: those based on compressive sensing (CS) \cite{bockelmann2013compressive,monsees2015compressive,gao2015compressive,du2017efficient,liu2018massive,senel2018grant,senel2017device,donoho2009message} covariance analysis (CV) \cite{HaghiJungCaire2018,chen2019covariance,ganesan2020clustering,Ganesan2020Algorithm}  and data-based approaches \cite{deSouza2023DeepLearning,bai2018deep}. Detecting active users within the GF-RA  protocol in figure \ref{fig1}(c) poses significant challenges, highlighting the need for advanced detection strategies. Grant-free access is a communication protocol that allows devices to transmit without prior scheduling permissions from the network controller, advantageous for environments requiring low latency and minimal signaling overhead, such as mMTC applications. In CF-mMIMO systems illustrated in figure \ref{fig1}(a), the stability provided by coherent blocks is critical for efficiently decoding unscheduled transmissions from multiple devices enabled by grant-free access, thereby enhancing network performance and scalability.

\subsection{Literature Review}
\noindent Hereafter, we provide a brief overview of the literature on sparse support recovery relevant to GF-RA protocols. The activity detection problem is formulated as a compressive sensing problem in \cite{bockelmann2013compressive,monsees2015compressive}, aiming to recover a sparse signal from a limited set of measurements. In GF-RA, these measurements correspond to the received signal at a given base station, while the sparse signal represents the sporadic activity patterns of the MTC devices. 

Several algorithms from the CS domain are used to detect device activity to jointly estimate the  channel. Greedy pursuit algorithms, commonly employed for sparse signal recovery \cite{gao2015compressive}, \cite{du2017efficient}, and the approximate message passing (AMP) algorithm \cite{liu2018massive} are notable examples. For instance, a non-coherent scheme proposed in \cite{senel2018grant} integrates information bits into devices' preamble sequences, facilitating joint activity and data detection. Despite the promising results of the AMP algorithm, challenges such as reliance on statistical channel information and stability issues prevent their practical implementations. CS-based algorithms exhibit significant performance degradation when the number of active devices exceeds the pilot sequence length \cite{senel2017device,donoho2009message}.

Covariance-based method is presented in \cite{HaghiJungCaire2018}, showing better performance than current CS-based techniques. The number of active devices should be less than the length of the pilot sequence, which is a constraint of  CS techniques that is addressed by the covariance-based one. Moreover,\cite{chen2019covariance} suggests a covariance-based strategy for combined activity and data detection, along with a performance analysis showing its advantages over AMP-based alternatives. 

\indent The authors in \cite{ganesan2020clustering} make several important contributions to the field of improving mMTC in CF-mMIMO  networks. They investigate GF-RA and use a covariance-based methodology to modeled the device activity detection as a maximum likelihood detection problem. Their approach, initially introduced in \cite{Ganesan2020Algorithm}, focuses on device activity detection via a single dominating access point (AP)  marking a noteworthy advancement. In addition, authors propose a novel clustering-based activity detection technique that promises to enhance detection efficiency at the expense of higher computation complexity on the central processing unit (CPU). Their results highlight how cell-free massive MIMO networks can greatly improve activity identification when a large number of devices are connected, demonstrating the potential of this method in future wireless networks aiming to facilitate IoT services.

\indent To greatly improve device activity identification in mMTC under GF-RA protocols, authors of \cite{deSouza2023DeepLearning} suggest a dual-strategy method. Their work aims at robust and efficient activity detection by introducing two deep learning (DL) algorithms for sparse support recovery. They include a feature selection step, which focuses on selecting various preamble sequences with advantageous correlation qualities, to optimize the neural network design. Their results show how effective Zadoff-Chu (ZC) sequences are as preambles, exhibiting higher activity detection accuracy with less computing load and providing a viable way to handle massive connectivity issues.

\indent To overcome the computational complexity of CS Multi-user Detection (CS-MUD) in huge mMTC systems,  in \cite{bai2018deep} authors present an innovative deep learning technique utilizing a specially designed block-restricted activation nonlinear unit, tailored to efficiently leverage the block sparse structure inherent in systems with many antennas or wide bandwidths. This approach enhances the performance of CS-MUD by offering a ten-fold reduction in computing time.
\subsection{Contributions}
\noindent Motivated by the significant performance improvements achieved with the data-based approach for MUD in \cite{deSouza2023DeepLearning}, where, an algorithm for activity detection in the context of GF-RA in mMIMO networks is developed. The following highlights the key contributions made by the present paper:

\begin{itemize}  
    \item To enable accurate device activity identification in mMTC GF-RA protocols, we present a robust and effective deep learning- based technique targeted at sparse support recovery. By utilizing data-driven neural networks, the DL-based model performs without requiring statistical information about the channel or device activity. The architecture employed is a deep multi-layer perceptron (DMLP), structured with densely connected layers, processing the symbols received during the random access slot efficiently.
    
    \item Numerical analysis is conducted on the architectural parameters of the DNN  to achieve a high degree of accuracy in detecting  activities as well minimizing the risk of overfitting.
     
    \item An analysis is undertaken to assess the performance of various activity detection scenarios and algorithms in CF-mMIMO systems. For instance, two different algorithms are thoroughly evaluated in a range of scenarios (e.g., dense AP deployment and with different cell area-size consideration, power-limited mMTC). Our analysis reveals a lower outage probability in our proposed scenario compared to those previously  in \cite{ganesan2020clustering}.
    
    \item Acknowledging the computational challenges of engaging all APs for user activity detection in a CF network, our research proposes a novel clustering majority-based method. This method optimizes the detection process by concentrating on a specific cluster of APs, rather than the entire network. This strategic approach not only simplifies the detection process but also ensures the preservation of the accuracy of the detection outcome as well as its integrity. This new clustering method decouples the clustering process from the MUD algorithm.
    
    \item We assess the effects of critical poor channel estimations on the performance of a mathematical model that has been referenced in  \cite{ganesan2020clustering}, emphasizing the vulnerability of such model-based techniques, resulting in a drop in performance as an immediate consequence of insufficient channel estimation.
    
    \item Within the digital signal processing domain, we assess the crucial role of converting from floating-point to fixed-point representation and its impact on the effectiveness of the computations, highlighting the significance of this conversion in preserving the operational performance of the proposed algorithm.
\end{itemize}

\indent The rest of this paper will be organized as follows: The signal model is explored in Section~\ref{sec:SystemModel}. Section~\ref{sec:DLAlgo} describes the DL-based algorithm in detail. Section~\ref{sec:Comparative_analysis} presents a comparative analysis of scenarios and algorithm applicability. The numerical results are provided in Section~\ref{sec:SimulationResults}, which show how effective the suggested algorithms are. Concluding remarks are included in Section~\ref{sec:Conclusion}.

\section{System Model}
\label{sec:SystemModel}
\noindent In the proposed system model, we investigate a CF-mMIMO wireless network composed of $M$ arbitrarily (or uniformly) positioned APs, each equipped with $N$ antennas, serving a set of $K$ users equipment (UE) with a single-antenna configuration. All $M$ APs are seamlessly linked to a CPU. Due to the sporadic nature of the traffic in the massive access scenario of mMTC, only a small fraction of the $K$ users are active at any given instant. \\
\indent In the sequel to this paper, we assume that each device independently transmits data based on a given activation probability, $\epsilon \ll 1$. We denote the device activity with $a_k \in \{0, 1\}$, where $a_k = 1$ denotes that the $k$-th device is active and $a_k = 0$ indicates silent device. The probability of activation $\Pr(a_k = 1)$ is denoted by $\epsilon$, and $\Pr(a_k = 0)$ is $(1-\epsilon)$. The resulting vector $\mathbf{a} = (a_1, a_2, \ldots, a_K)$ denotes the activity of $K$ users at any time instant, and due to the sporadic nature of the traffic, this vector is sparse resulting from the small value of $\epsilon$. The set of active users is represented by $\mathcal{A}$, where $\mathcal{A}=\{ k : a_k=1\}$. Additionally, it is noteworthy that the activity follows the Bernoulli distribution.\\
\indent Considering the geographical separation of all APs and devices, we assume independent channels between devices and APs. Additionally, we assume that the $N$ antennas at each AP are sufficiently separated to exhibit independent fading. The channel gain, denoted as $\gmkn$, between the $n\textsuperscript{th}$ antenna in the $m\textsuperscript{th}$ AP to device $k$ is given by:

\begin{equation}
\label{eq:Channelfad}
\gmkn = \beta_{mk}^{1/2} \times \hmkn,
\end{equation}
where $\beta_{mk}$  is defined as the large-scale fading coefficient between the $m\textsuperscript{th}$ AP and user $k$ as specified in the 3GPP standard \cite{3gppTR38901} given by :
 \begin{align}
\beta_{mk} &= PL + F_{mk} \notag \\
PL &= 
\begin{cases} 
PL_1, & 10m \leq d_{2D} \leq d'_{BP} \\
PL_2, & d'_{BP} < d_{2D} \leq 5km 
\end{cases} \notag \\
PL_1 &= 28.0 + 22\log_{10}(d_{3D}) + 20\log_{10}(f_c) \notag \\
PL_2 &= 28.0 + 40\log_{10}(d_{3D}) + 20\log_{10}(f_c) \notag \\
&\quad - 9\log_{10}\left(\left(d'_{BP}\right)^2 + (h_{BS} - h_{UT})^2\right) , \label{eq:largescale}
\end{align}
the large-scale fading coefficient \(\beta_{mk}\) consists of two elements: the standard path loss model \(PL\), pertinent to urban macro line-of-sight (UMa-LOS) conditions, and the shadow fading component modeled as a gaussian random variable as \(F_{mk} \sim \mathcal{N}(0,\sigma_{sh}^2)\). The path loss model is contingent on the horizontal distance \(d_{2D}\), gauging the separation between each user \(k\) and access point \(m\), with \(d'_{BP}\) denoting the breakpoint distance beyond which the model switches from \(PL_1\) to \(PL_2\), accounting for the height differences between user and AP.
Furthermore, the small-scale fading coefficient is denoted by \( \hmkn \sim \mathcal{CN}(0,1) \) and remains constant for some consecutive activity vectors, where the activity can change up to multiple times before the channel changes. The system operates in a block-fading scenario where each channel remains constant during a coherence time interval, and all channels are independently distributed.\\
\indent In the uplink of a narrow-band mMTC system within a square area of $D \times D$ km\textsuperscript{2}, assigning orthogonal pilot sequences to users is unfeasible due to limited channel coherence time $\tau_C$  defined as the duration over which the channel remains approximately constant. The system typically has a large number of users, denoted as $K$, which significantly exceeds $T_C$. Here, $T_C$ represents the number of symbols that can be transmitted in the coherence block as illustrated in figure \ref{fig1}(b) and is derived as $T_C = \tau_C \times B_C$, with $B_C$ as the coherence bandwidth. The phenomenon of pilot contamination arises because each active device transmits non-orthogonal pilot sequences $\mathbf{s_k}$ during a random access slot. These sequences are drawn from a Gaussian distribution, i.e., $\mathbf{s_k} \in \mathbb{C}^{L\times 1}$, where $L$ is the pilot sequence length. The length $L$ constitutes a specific reserved portion, such as $\Delta$ percentage, of the coherence block. All frames are assumed to be received synchronously at the BS.\\
The received signal $\mathbf{y_{mn}} \in \mathbb{C}^{L\times 1}$ at the $n$\textsuperscript{th} antenna of the $m$\textsuperscript{th} AP is expressed as :   
\begin{equation}
\label{eq:SignalrecN}
\begin{aligned}
\mathbf{y_{mn}} &= \sum_{k=1}^{K} a_k \rho_k^{1/2} \gmkn s_k + w_{mn} \\
&= \mathbf{SD_a D_{\rho}}^{1/2} \mathbf{ g_{mn}} + \mathbf{w_{mn}}, 
\end{aligned}
\end{equation}
where $\mathbf{S} = [\mathbf{s_1} \; \mathbf{s_2} \; \ldots \; \mathbf{s_K}] \in \mathbb{C}^{L \times K}$
is the collection of all pilot sequences,  $\rho_k$ is the power transmitted by user $k$ where the matrix $\mathbf{D_{\mathbf{\rho}}} = \text{diag}(\rho_1, \rho_2, \ldots, \rho_K)$, $\mathbf{D_a} = \text{diag}(\mathbf{a})$ is the activity matrix ,  $\mathbf{g_{mn}} = [g_{m1}^{(n)} \; g_{m2}^{(n)} \; \ldots \; g_{mK}^{(n)}] \in \mathbb{C}^{K \times 1}$ is the channel vector from all $K$ users to the $n\textsuperscript{th}$ antenna of the $m\textsuperscript{th}$ AP, and $\mathbf{w_{mn}} \sim \mathcal{CN}(0,\sigma^2 I_L)$ is the independent additive white gaussian noise vector.\\
Thus, the signal $\mathbf{Y_{m}} \in \mathbb{C}^{L \times N}$  received at the $m\textsuperscript{th}$ AP can be expressed as:
\begin{equation}
\begin{aligned}
\label{eq:SigRecM}
\mathbf{Y_{m}} &= \mathbf{SD_aD_{\rho}}^{1/2}\mathbf{G_m} + \mathbf{W_m}, \\
\end{aligned}
\end{equation}
where $\mathbf{G_m} = [\mathbf{g_{m1}} \; \mathbf{g_{m2}} \; \ldots \; \mathbf{g_{mN}}] \in \mathbb{C}^{K \times N}$ is the channel matrix between the $K$ users and the $m\textsuperscript{th}$ AP, and $\mathbf{W_m} = [\mathbf{w_{m1}} \; \mathbf{w_{m2}} \; \ldots \; \mathbf{w_{mN}}] \in \mathbb{C}^{L \times N}$ is the noise matrix.\\
In the subsequent analysis, we will omit the index $m$ from (\ref{eq:SigRecM}), which is defined as :
\begin{equation}
\begin{aligned}
\label{eq:SigRec}
\mathbf{Y} = \mathbf{D_aB} +\mathbf{W},
\end{aligned}
\end{equation}
where  $\mathbf{B} = \mathbf{SD_{\rho}}^{1/2}\mathbf{G_m} $ and $\mathbf{W} = \mathbf{W_m} $. The device activity detection problem aims to find the most likely activity vector given the code book of pilot sequences $\mathbf{S}$. This is equivalent to identifying the indices of the nonzero entries in the sparse vector $\mathbf{a}$ that corresponds to the diagonal of \(\mathbf{D_a}\) , which is the vector that represents $\mathbf{a}$ support :
\begin{equation}
\label{eq:Support}
\text{supp}(\mathbf{a}) = \{ k \in \mathbb{N}^+ \, | \, a_k \neq 0 \}.
\end{equation}

Determining the support of $\mathbf{a}$ from $\mathbf{Y}$ and $\mathbf{S}$ can be expressed as a sparse support recovery problem given the condition of sporadic activation. The user activity detection (UAD)  problem can then be formulated as follows:
\begin{equation}
\label{eq:ProbReformulation}
P_0 : \min_{\mathbf{X}} \left \| \mathbf{Y} - \mathbf{D_aB} \right \|_F ,
\end{equation}
where, $\|\cdot\|_{(F)}$ denotes the Frobenius norm, which is appropriate for matrix data and simplifies to the sum of the absolute squares of the differences between corresponding elements of \(\mathbf{Y}\) and \(\mathbf{D_aB}\). In the compressive sensing theory, the optimization problem \(P_{0}\) is an NP-hard problem with a non-polynomial number of combinations. Common approximations, such as convex relaxation methods and iterative algorithms, as well as various other approaches, offer varying degrees of computational efficiency to tackle the \(P_{0}\) computation complexity.\\
\indent The optimization problem \(P_{0}\) can be addressed using specialized tools, such as DL techniques for mMTC, particularly for detecting device activity and recovering the sparse support vector.
\begin{figure}[h]
  \centering
  \includegraphics[width=\columnwidth]{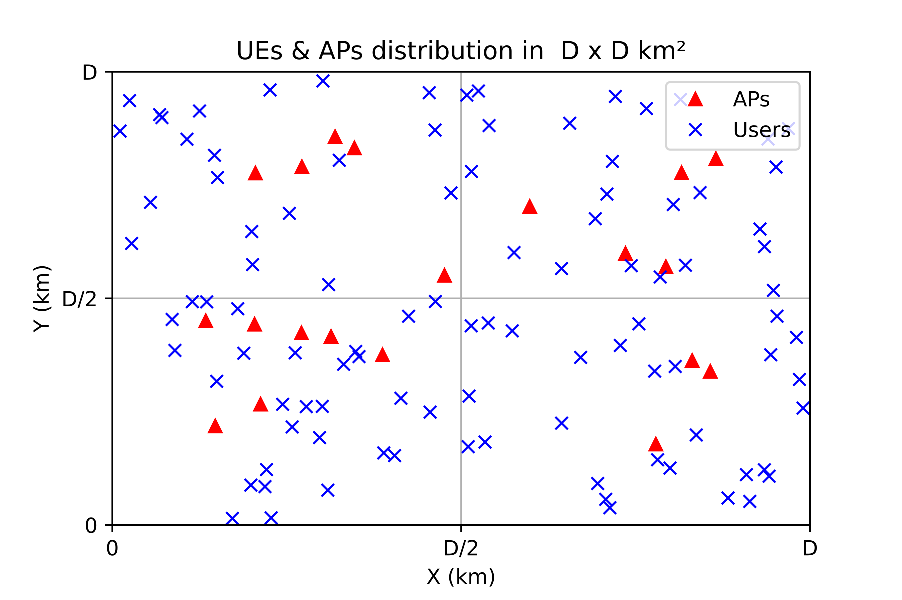}
  \vspace{-10pt} 
  \caption{Distribution of users and access points within the network.}
  \label{fig2}
\end{figure}
\section{Deep Learning Algorithm}
\label{sec:DLAlgo}
\noindent Regression analysis using a Deep Neural Network (DNN) architecture involves mapping an input signal into a specific output. A more capacity-efficient approach is prompted by realizing that a sparse vector can be efficiently derived by the least squares estimator, particularly when its support is known. We suggest using a DNN to approximate the mapping from $\mathbf{Y}$ to the support of the vector $\textbf{a}$ in (\ref{eq:Support}) rather than directly predicting $\textbf{a}$. This intentional decision turns the issue into a multi-binary label classification challenge, which is very useful for identifying active users.

\indent This section presents an overview of the process of data generation, training phase and introduces the architecture of the DMLP, which is specifically made for UAD and uses a variety of input features to efficiently detect activity.

\subsection{Data Generation}
\noindent The challenge of gathering large amounts of real data is a major problem when using DL in wireless communication. Fortunately, synthetic data can train DNNs to tackle optimization challenges in (\ref{eq:ProbReformulation}). Our research scenario on CF-mMIMO systems took into account the need to accurately simulate real-world environments during the data production phase. \\
\indent At first, this involved allocating $K$ users to a designated surface area of $D \times D \,  \text{km}^2$, with $M$ APs, each equipped with $N$ antenna components, as depicted in figure \ref{fig2}. By maintaining the AP-AP and UE-AP distances between each other at a minimum, this provided a realistic spacing, edge distance, and coverage distribution.\\
\indent As the geographic topology was set up, the large-scale fading effects were calculated using the 3GPP standard UMa criteria, incorporating the LOS condition. This was important because it highlighted how complex signal transmission is in  metropolitan environments, where massive structures (distance attenuation, shadowing) have a significant impact on received signal behavior. \\
\indent Furthermore, each user likelihood of being active was indicated by their activity status, calculated with a binomial distribution. We aimed for 95\% of active devices in our study to achieve the lowest signal-to-noise ratio (SNR) for network access. This involved ensuring that all active users attained the target SNR at their dominant AP, which is chosen based on the maximum large-scale fading factor $\beta_{mk}$ between users and access points.\\
\indent Two strategies are available for modeling rayleigh fading channels. The first permits fading to change with user activities, accurately reflecting real conditions and evaluating the system's response to variations. The second strategy employs a consistent fading model across several activities. Utilizing these approaches enables a detailed examination of system behavior in both dynamic and stable scenarios, yielding valuable insights into its overall performance.\\
\indent Finally, the received signal at each AP was calculated using equation (\ref{eq:SignalrecN}), which combined transmitted pilot sequence, large- and small-scale fading effects, and gaussian noise. This comprehensive approach to data collection offered a robust dataset, which was essential for evaluating the performance of the proposed algorithm in simulated yet close-to-realistic CF-mMIMO network scenarios.
\protect\subsection{Training Phase }
\noindent In practical settings, our model operates initially in two distinct phases due to the absence of available training data. The initial phase, termed the transitional phase, entails the accumulation of the required data for model training. Subsequently, our research predominantly concentrates on the second phase, referred to as the permanent phase, during which the model uses an established dataset. The simulations and results of this study are rooted in the permanent phase, assuming that the model has progressed beyond the preliminary stage of data collection.
\indent Let  \( D_{tr} = \{ (\mathbf{Y_i, a_i}) \}_i \) where \( I = |D_{tr}| = 3 \times 10^4 \) be a training dataset containing the received signal and the activity descriptor of a random access slot. Once the data are produced then we can train the model, the optimization problem can be stated as follows:
\begin{align}
g(Y, S, \theta) &= \min_{\mathbf{X}} \left \| \mathbf{Y} - \mathbf{D_aB} \right \|_F,   \label{eq:combined_optimization}
\end{align}
where the model parameters are indicated by \( \theta \). A DNN is trained to map each input \( Y_i \) to a desired estimated activity indicator  \( \mathbf{a}_i \) via several successive layers of linear transformation interspersed with element-wise nonlinear transforms, given a collection of training examples \( D_{tr} \). The input-output relationship for the hidden layers in a standard feed-forward neural network (FFNN) is expressed as follows:
\begin{equation}
\label{eq:In_Out_Relation_FFNN}
\mathbf{x^{(t)}} = f\left(\mathbf{U_{t}} \mathbf{x^{(t-1)}} + \mathbf{b_{t}}\right), 
\end{equation}

where \( f(\cdot) \) denotes the non-linear activation function. In this equation, \( \mathbf{x^{(t)}} \) represents the output of the layer (t), with \( \mathbf{x^{(t-1)}} \) being the input received from the previous layer (t-1). The parameters of the network at this stage, namely the weight matrix \( \mathbf{U_{t}} \) between layer $t-1$  and layer $t$, and the bias vector \( \mathbf{b_{t}} \) at layer $t$, are collectively represented by \( \theta \).
\indent These parameters are learned and updated during training, utilizing the Adam optimizer. This optimizer, a popular variant of stochastic gradient descent, dynamically adjusts the learning rate for each parameter. The Adam optimizer is particularly popular in machine learning (ML) due to its adaptability to handle sparse gradients and use of an adaptive learning rate mechanism, which allows it to converge in complicated scenarios with resilience. We denote the output of the DL algorithm for the input sample \( (\mathbf{Y_i}, \mathbf{a}_i) \) as \(\mathbf{\tilde{a}}_i \). For the network training, we use the binary cross-entropy loss\cite{Nam2014LargeScale} calculated over all the \(I = |D_{tr}|\) samples:

\begin{equation}
\label{eq:BinCrossEntro}
\mathcal{L}(D_{tr}) = -\sum_{i = 1}^{I} \left[ \mathbf{a_i} \log(\mathbf{\tilde{a}_i}) + (1 - \mathbf{a_i}) \log(1 - \mathbf{\tilde{a}_i}) \right] 
\end{equation}

\indent Using the back-propagation technique, we may propagate the loss all the way and train the parameters of the network. The loss function \(\mathcal{L}(D_{tr})\) is minimized during the training process by choosing the right parameters \(\theta\). If after multiple epochs the loss function in (\ref{eq:BinCrossEntro}) does not improve, the training ends. That is, the trained DL network estimates the activity descriptor with high accuracy.
\indent There is an offline option for the training procedure. After the DNN parameters are learned, it is computationally less expensive to use this neural network for identifying active users in new datasets (inference phase) because it only requires a few vector-matrix multiplications and summations as well as element-wise nonlinear operations.

\begin{figure}[h]
  \centering
  \includegraphics[width=\columnwidth]{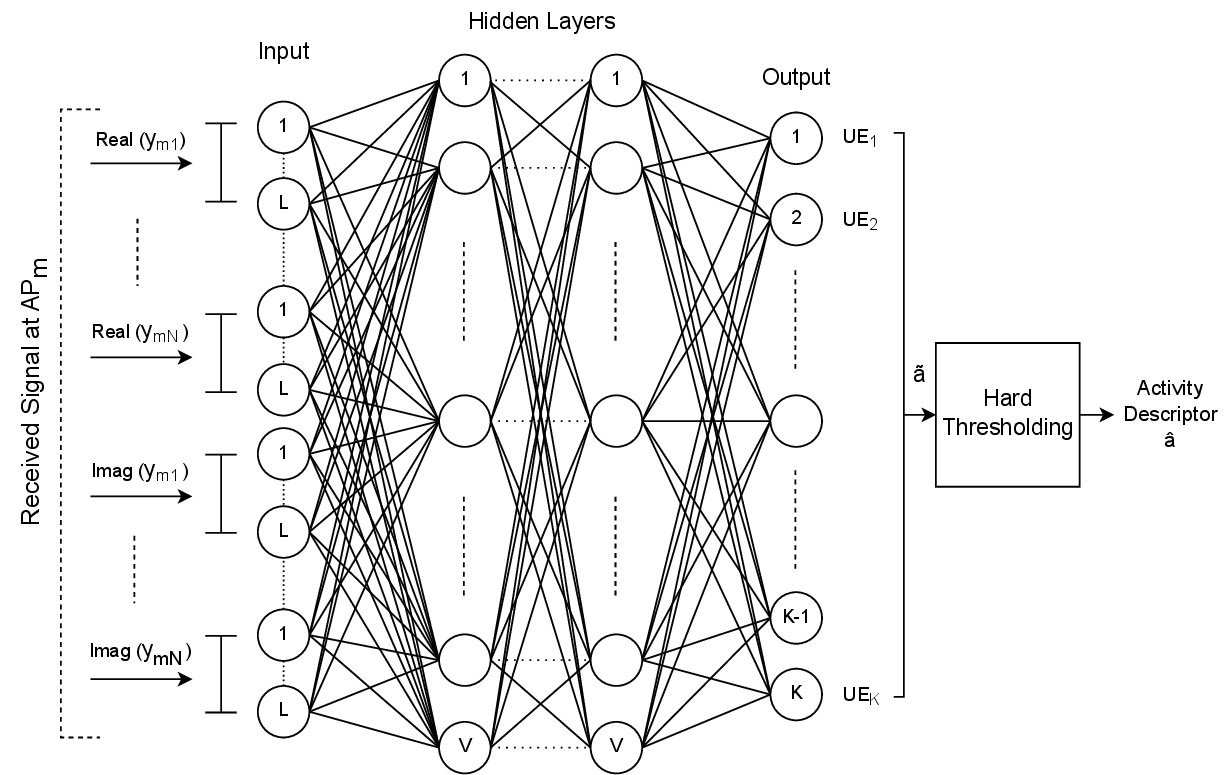}
  \vspace{0.5pt} 
  \caption{Architecture of the DMLP algorithm proposed for UAD.}
  \vspace{-20pt} 
  \label{fig3}
\end{figure}

\protect\subsection{Deep Multilayer Perceptron (DMLP)}
An extremely popular neural network form is the Deep Multilayer Perceptron (DMLP). The architecture is depicted in figure \ref{fig3} and consists of an input layer that processes incoming data, such as the real and imaginary components of signals at each antenna element  $n$  of the access point  $AP_m$  as $\mathbf{Y_m} = [\mathbf{y_{m1}} \;  \mathbf{y_{m2}} \; \ldots \; \mathbf{y_{mN}}] \in \mathbb{C}^{L \times N} $. This architecture also includes an output layer, equipped with $ K $ neurons, for predicting and outputting the activity descriptor vector, as well as $Z$  intermediate hidden layers densely connected. Each of these layers contains several nodes,  $V$ , which establish forward connections to the neurons of the subsequent layer.

\indent Based on (\ref{eq:In_Out_Relation_FFNN}), the hidden layers in the model employ the ReLU (Rectified Linear Unit) as their activation function because it is widely used in the hidden layers of FFNN due to its ability to introduce non-linearity while being computationally efficient, as it supports the use of back propagation for efficient learning in complex and deep network architectures:
\begin{equation}
\label{eq:ReLU}
\textit{ReLU}(c) = \max(0, c). 
\end{equation}
\indent Besides, the sigmoid activation function is employed in the output layer because it plays a crucial role in binary classification tasks. This decision is based on the particular ability to perform binary classification for each device in order to identify their activities, and it is provided by: 
\begin{equation}
\text{Sigmoid}(c) = \frac{1}{1 + e^{-c}}. 
\end{equation}
\indent The input and output dimensions of the DMLP are mainly determined by three key criteria: the pilot sequence  length $L$, the number of antenna elements 
$N$ at each access point, and the total number of devices $K$. Consequently, any changes in the number of devices or the duration of the preamble require retraining for the network to effectively adapt to these new parameters.

\indent The weight matrices and bias vectors, which are the parameters of the DMLP layers, are now introduced. We define the weight matrices according to the dimensions of the inputs and outputs of each layer as follows:
\( \mathbf{U_1} \in \mathbb{R}^{V \times 2NL} \)  (weight matrix between the input layer and the first hidden layer), \( \mathbf{U_2} \in \mathbb{R}^{V \times V} \), \(\dots\) , \( \mathbf{U_Z} \in \mathbb{R}^{V \times V} \) and \( \mathbf{U_{Z+1}} \in \mathbb{R}^{K \times V} \) for the hidden layers $1$
, $2$, \(\dots\) , $Z$, and the output layer ($Z+1$), respectively. In the same way, the bias-valued vectors are \( \mathbf{b_1} \in \mathbb{R}^{V \times 1} \), \( \mathbf{b_2} \in \mathbb{R}^{V \times 1} \) , \(\dots\) , \( \mathbf{b_Z} \in \mathbb{R}^{V \times 1} \) and \( \mathbf{b_{Z+1}} \in \mathbb{R}^{K \times 1} \). Considering this, we use the DMLP parameters to define the set by:
\begin{equation}
\label{eq:DMLP_Para}
\theta_{\text{DMLP}} = \{ \mathbf{U_t}, \mathbf{b_t} \}, \quad t = 1, 2, 3, \ldots 
\end{equation}

\indent Finding the number of trainable parameters in the DMLP method is feasible when one knows $\theta_{\text{DMLP}}$. The number of trainable parameters is determined by taking into account the values of the bias vectors and the weights matrices:
\begin{equation}
\label{eq:DMLP_Parameter}
\theta_{\text{DMLP}}(K, L, N, V, Z) = (Z-1)V^2 + (2NL + K + Z)V + K.
\end{equation}

\indent A hard decision mechanism with a threshold parameter $\tau \geq 0$ is set up at the output of the algorithm to calculate the activity descriptor in its original domain, as the outputs $\mathbf{\tilde{a}}$ of the DMLP are probability values that fall within the range \([0, 1]\). Thus, the estimated activity indicator for each device (the output of the hard decision mechanism) is given by:
\begin {equation}
\hat{a}_k = 
\begin{cases}
0, & \text{if } \tilde{a}_k < \tau \\
1, & \text{if } \tilde{a}_k \geq \tau
\end{cases} 
\end{equation}

\indent In the process of detecting device activity, two primary types of errors can occur. The first is a false alarm (FA), which happens when a device that is not active is incorrectly identified as active. The second type is a miss detection (MD), occurring when an active device is wrongly classified as inactive. The likelihoods of both FA and MD are determined using a specific hard decision threshold.
\begin{equation}
\label{eq:Prob_FA}
P_{\text{FA}}(\tau) = \Pr(\tilde{a}_k > \tau \,|\, a_k = 0) 
\end{equation}
\begin{equation}
\label{eq:Prob_MD}
P_{\text{MD}}(\tau) = \Pr(\tilde{a}_k < \tau \,|\, a_k = 1) 
\end{equation}

\indent The probability of error within each random access slot is affected by the varying number of active devices, which in turn alters the frequency of specific error types. This variability is encapsulated in the error probability function that is expressed using the false alarm and miss detection probabilities related to the threshold \(\tau\) and the probability of activation \(\varepsilon\) as:
\begin{equation}
\label{eq:Probability_Error}
P_E(\tau, \varepsilon) = (1 - \varepsilon) P_{\text{FA}}(\tau) + \varepsilon P_{\text{MD}}(\tau)
\end{equation}

\indent The optimization of the hard decision module can be tailored to satisfy various design objectives. For instance, it might be configured to minimize a particular error type or to balance different errors by applying distinct weights to them. In our research, we approach the calibration of the hard decision threshold in a way that for a specific \(\varepsilon = \varepsilon_0\):
\begin{equation}
\tau^* = \min_{\tau \geq 0} P_E(\tau, \varepsilon_0). 
\end{equation}

\indent Due to the complexity of deriving an exact formula for the DL algorithm output error, we instead approximate the near-optimal hard decision threshold $\tau$ (notated as $\tilde{\tau}^*$) through numerical methods as in figure \ref{fig8}. This involves adjusting $\tau$ based on training data samples to align with a specific criterion, using predictions from our trained NN. The most effective threshold is identified by examining the recovery rate of samples across the predicted output vectors.

\protect\subsection{Clustering-Based Activity Detection}
\noindent The traditional activity detection algorithm depends on data gathered from just one access point for each device. However, in a CF-mMIMO network, the most effective approach would involve all AP contributing to activity detection for all users. Nonetheless, implementing such a method proves to have excessive computation complexity. An alternative is clustering-based activity detection algorithms for GF-RA, as studied in \cite{ganesan2020clustering}, where groups AP with strong channels to individual users. This method, while efficient, involves knowing the large-scale fading \(\beta_{mk}\), iterating over all users over different clusters. It leads to a considerable computing effort as the number of users and cluster sizes $T$ increase. Here, $T$ represents the number of APs per cluster. This requires solving a polynomial equation of degree \(2T-1\) to find the roots \cite{ganesan2020clustering}.

\indent Our methodology differs from the previously stated method, as we have devised a novel strategy while retaining the cluster-based framework. We streamline the process by foregoing the selection of the optimal cluster per user. Instead, we adopt a more time-efficient approach by randomly selecting clusters for all users simultaneously. This dynamic selection process, which can evolve over time in response to user activity patterns, enables us to detect activity for all users in a single step. Consequently, this enhances flexibility and efficiency in activity detection within cell-free networks, resulting in a substantial reduction in computational complexity.

\indent Each AP in the selected cluster receives data from all active users, processes it using the DMLP algorithm, followed by a hard decision phase, and sends its results to a CPU via the Fronthaul link (see figure \ref{fig1}(a)). The CPU then applies a majority decision rule to determine the final activity status of each user. This decision depends on whether the number of APs in the cluster is even or odd. In even cases, a user is considered active if at least half of the APs (i.e., $T/2$ APs) in the cluster indicate activity. In odd cases, the strict majority (i.e., $(T+1)/2$  APs) determines the activity status.
\indent The suggested approach is distinctive because it processes user activity detection comprehensively in a single step, free from iterative procedures, using a randomly selected cluster of access points. Additionally, it retains its effectiveness even as cluster size expands, without the necessity to solve polynomial equations of higher degrees. Additionally, it eliminates the need to rely on large-scale fading data for cluster selection.
\begin{table}[h]
\centering
\caption{Simulation Parameters.}
\label{table:Parameters}
\begin{tabularx}{\columnwidth}{|X|l|c|c|}
\hline
\multicolumn{2}{|c|}{\textbf{Parameter}} & \textbf{Scenario I} & \textbf{Scenario II} \\
\hline
\multicolumn{4}{|c|}{\textbf{System}} \\
\hline
\multicolumn{2}{|l|}{Area Size (D\textsuperscript{2})} & 0.25 Km\textsuperscript{2} & 1 Km\textsuperscript{2} \\
\hline
\multicolumn{2}{|l|}{Edge Distance} & 50 m & - \\
\hline
\multicolumn{2}{|l|}{Number of Users (K)} & 100 & 100 \\
\hline
\multicolumn{2}{|l|}{Number of APs (M)} & 20 & 20 \\
\hline
\multicolumn{2}{|l|}{Number of Antennas (N)} & 2 & 2 \\
\hline
\multicolumn{2}{|l|}{Pilot Length (L)} & 40 & 40 \\
\hline
\multicolumn{2}{|l|}{Minimum distance UE-AP} & 10 m & - \\
\hline
\multicolumn{2}{|l|}{AP Spacing} & 15 m & - \\
\hline
\multicolumn{2}{|l|}{Antenna Height} & 12 m & - \\
\hline
\multicolumn{2}{|l|}{User Height} & 1.5 m & - \\
\hline
\multicolumn{2}{|l|}{Carrier Frequency (fc)} & 900 MHz & - \\
\hline
\multicolumn{2}{|l|}{Cluster Size T} & 1,2,3,... & 1,2,3,... \\
\hline
\multicolumn{2}{|l|}{Transmit Power (TxPow)} & 200 mW & 200 mW \\
\hline
\multicolumn{2}{|l|}{Activity Probability (\(\varepsilon\))} & 0.1 & 0.1 \\
\hline
\multicolumn{4}{|c|}{\textbf{Coherence Block}} \\
\hline
\multicolumn{2}{|l|}{Coherence Time \(\tau_c\)} & 1 ms & 1 ms \\
\hline
\multicolumn{2}{|l|}{Coherence Bandwidth \(B_c\)} & 200 kHz & 200 kHz \\
\hline
\multicolumn{2}{|l|}{Number of symbols \(T_c\)} & 200 & 200 \\
\hline
\multicolumn{2}{|l|}{Reserved portion \(\Delta\)} & 20\% & 20\% \\
\hline
\multicolumn{4}{|c|}{\textbf{Channel Fading}} \\
\hline
\multicolumn{2}{|l|}{Noise power \(\sigma^2\)} & -109 dBm & -109 dBm \\
\hline
\multicolumn{2}{|l|}{Small Scale Fading } & Rayleigh & Rayleigh \\
\hline
\multirow{2}{*}{\makecell{Large Scale\\Fading}} & Path Loss & 3GPP Standard & \multicolumn{1}{c|}{Simplified Model} \\
\cline{2-4}
 & Shadowing & \multicolumn{1}{c|}{\(\mathcal{N}(0,1)\)} & \(\mathcal{N}(0,4)\) \\
\hline
\end{tabularx}
\end{table}

\section{Comparative Analysis of Scenarios and Algorithms Applicability }
\label{sec:Comparative_analysis}
\noindent In our investigation, we meticulously assessed two distinct activity detection algorithms in cell-free massive MIMO systems across different scenarios.  Initially, we explored the mathematical covariance-based methodology from \cite{ganesan2020clustering}. Subsequently, we examined our data-driven approach, utilizing neural networks, specifically the Deep Multi-Layer Perceptron. \\
\indent Our primary scenario is thorough, featuring a range of comprehensive parameters, while the second scenario, adopted from another research work \cite{ganesan2020clustering}, is less detailed. However, it has been considered and tested to provide comparison elements with the results derived from the work that introduced it in \cite{ganesan2020clustering}. The key parameters for both scenarios are outlined in Table~\ref{table:Parameters}.
\footnote{See Section ~\ref{sec:SystemModel} for a comprehensive analysis of simulation assumptions and parameters. Briefly, with $\tau_C = 1$ ms and $B_C = 200$ kHz yielding $T_C = 200$ symbols, and given $K = 100$ users, orthogonal pilot sequences might be deemed feasible. However, In our simulations, we employ non-orthogonal pilot sequences, even though $K$ does not exceed $T_C$, to improve the timing efficiency of our simulations. Given our objective to expand connectivity, this strategy allows for a potential increase in the number of users. This approach is justified by its minimal effect on the outcomes.}

\vspace{-10pt}
\protect\subsection{Algorithm Applicability}
\noindent The methodological approach conducts a comprehensive analysis by applying both data-driven machine learning and the mathematical algorithm proposed in \cite{ganesan2020clustering} to these distinct scenarios. This section will provide a detailed description of the outcomes of the cross-application testing of each algorithm in different scenarios. Despite using similar metrics, receiver operating characteristics (ROC), which plots the true positive rate (TPR) against the false positive rate (FPR). A direct comparison of the algorithms is not feasible due to different modeling techniques, assumptions, and differences in their outputs.
\begin{figure}[h]
  \centering
  \includegraphics[width=\columnwidth]{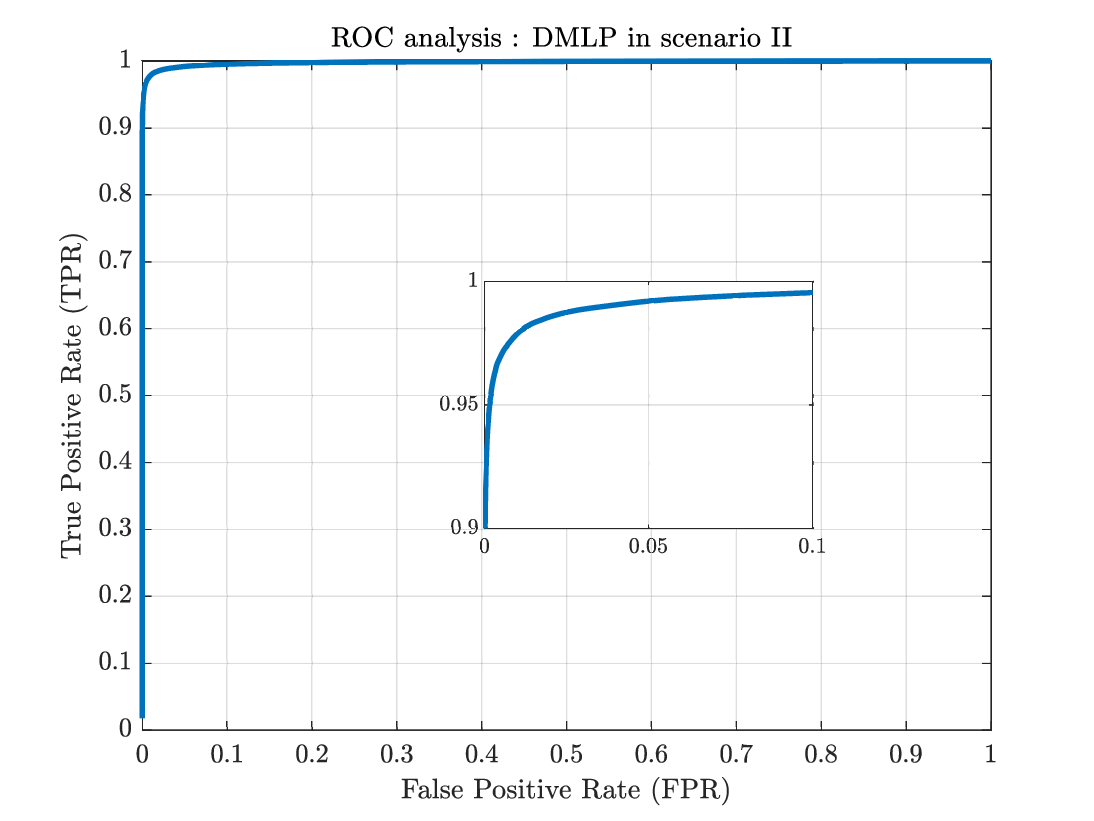}
  \vspace{-15pt} 
  \caption{DMLP algorithm performance in scenario II from \cite{ganesan2020clustering}.}
  \label{fig4}
\end{figure}

\indent Effective user activity detection is demonstrated by figure \ref{fig4}, which shows the ROC curve for the DMLP algorithm in scenario II  (Table~\ref{table:Parameters}). It shows a strong detection probability with few false alarms. The ability of the algorithm to distinguish between active and non-active users is highlighted by its steep ascent towards the upper left corner, indicating its effectiveness in the present scenario.
\begin{figure}[!t]
  \centering
  \includegraphics[width=\columnwidth]{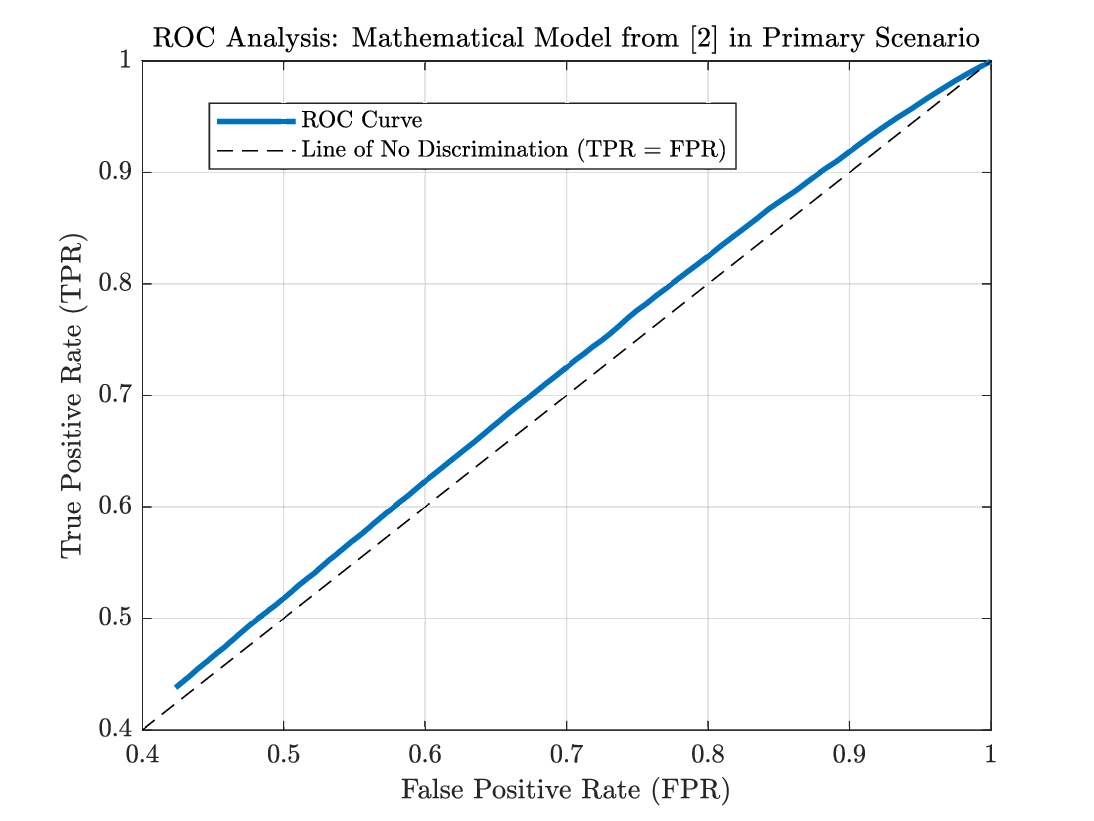}
  \vspace{-15pt} 
  \caption{Mathematical approach performance in scenario I.}
  \vspace{-15pt}
  \label{fig5}
\end{figure}

\indent As presented in \cite{ganesan2020clustering}, the proposed model has demonstrated excellent results in its scenario II. We have successfully regenerated the results shown in the paper, confirming their validity and consistency. However, while employing the same model in \cite{ganesan2020clustering} to scenario I in (Table~\ref{table:Parameters}), the performance differs significantly. The ROC curve shown in figure \ref{fig5} highlights a performance that is near the line of no discrimination, indicating a low capacity for prediction. TPR match FPR across the spectrum, indicating that there is little difference between active and inactive users. This result emphasizes the minimal distinction between  user activity states.

\indent Furthermore, the channel fading effects represent the primary difference between the two scenarios being compared. This variance plays a key role in the reported results, indicating that the performance of the algorithm in \cite{ganesan2020clustering} is very sensitive to variations in the channel fading circumstances. It also exposes the model  \cite{ganesan2020clustering} vulnerability and its inability to scale in comparison to other channel-fading models. This algorithm is conditioned by the scenario model.
\vspace{-5pt}
\protect\subsection{DMLP  Performance Summary}
\noindent After evaluating the cross-application of each algorithm across varied scenarios, it becomes evident that our DMLP algorithm maintains consistent performance, adapting seamlessly to the more elementary scenario from \cite{ganesan2020clustering}. Conversely, the reference algorithm, limited by its basic design, cannot extend its application in scenario I. 

\section{Numerical Results}
\label{sec:SimulationResults}
\noindent In this section, we explore the topic of improved connectivity in cell-free massive MIMO structures, assessing our suggested algorithm effectiveness in real-world circumstances and comparing them, when necessary, to specific mathematical approaches. The inherent challenges of directly evaluating the two models call for this comparative analysis, especially in light of the conclusions from the previous section. To accurately assess the efficiency of activity detection in these systems, our analysis utilizes two distinct metrics: the ROC curve, and the probability of miss detection (PMD) as a function of the probability of false alarm (PFA) presented in a log-log scale. This dual approach allows for a comprehensive evaluation of detection performance across different operational thresholds.

\begin{figure}[h]
  \centering
  \includegraphics[width=\columnwidth]{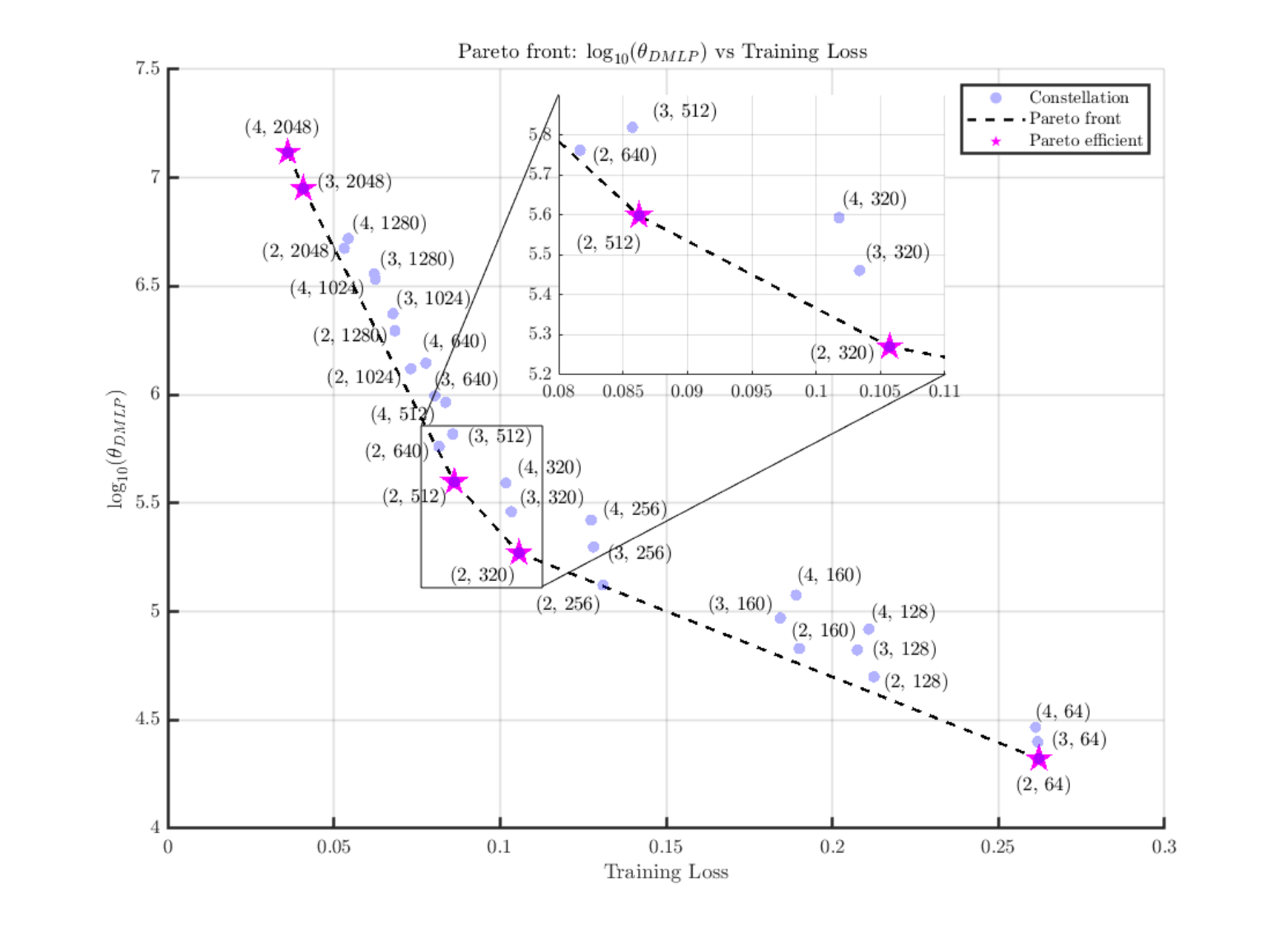}
  \vspace{-15pt} 
  \caption{Pareto front : number of trainable parameters in the DMLP/ training loss.}
  \label{fig6}
\end{figure}

\protect\subsection{DMLP Parameters Tuning}
\noindent In this section, our primary focus is to evaluate the performance of DNN in the context of user activity detection in CF-mMIMO systems. To achieve this goal, we have embarked on a comprehensive investigation into the architectural parameters of DNN. Our objective is to achieve a high level of accuracy in activity detection while mitigating the risk of over-fitting.

\indent The tuning process involves careful consideration of the number of neurons within the hidden layers of the DMLP architecture. We have conducted a thorough parameter search, varying the number of neurons $V$  within the range \{64, 128, 160, 256, 320, 512, 64, 1024, 1280, 2048\} and exploring different detector configurations with various numbers of intermediate hidden layers $Z$ within the set \{2, 3, 4\}. This meticulous parameter-tuning process is an essential step to optimize the DNN for accurate activity detection.

\indent To objectively study the performance-complexity trade-off, the detectors are compared based on Pareto efficiency. A detector (\(Z_i,V_j\)) is considered Pareto efficient if no alternative detector exists that reduces either the complexity, assessed as the number of training parameters as shown in (\ref{eq:DMLP_Parameter}), or the training losses using binary cross-entropy on the training dataset in (\ref{eq:BinCrossEntro}), without adversely affecting the other metric. The set of all Pareto-efficient detectors, known as the Pareto front, represents the available trade-off options. Switching from one Pareto-efficient detector to another involves prioritizing either complexity or training loss. Conversely, a detector that is not Pareto efficient should not be selected, as it is possible to improve at least one of the metrics without compromising the other.
 
\indent Analyzing the curve in figure ~\ref{fig6}, it is evident that increasing the number of parameters consistently decreases the minimum training loss. This trend is depicted in figure  \ref{fig6}, where it is noticeable that varying the number of layers results in only a slight improvement in the validation loss for each $N_n$ value. However, a substantial decrease in loss is observed when the number of neurons is increased for each $Z$ value. 

\indent For a more detailed comparison, we examine two points on the Pareto front: (2, 320) and (2, 512), as shown in figure \ref{fig6}. Although the detectors in the range of 0.1 loss value are close in terms of loss, choosing the detector with configuration (2, 512) requires twice the number of parameters to be trained compared to the (2, 320) configuration. This significant increase in complexity not only impacts the training time but also the overall efficiency of the detector due to the more complex architecture. Therefore, for the DMLP algorithm detectors, the best trade-off is achieved with $Z = 2$ and $V = 320$.

\vspace{-5pt}
\begin{figure}[h]
  \centering
  \includegraphics[width=\columnwidth]{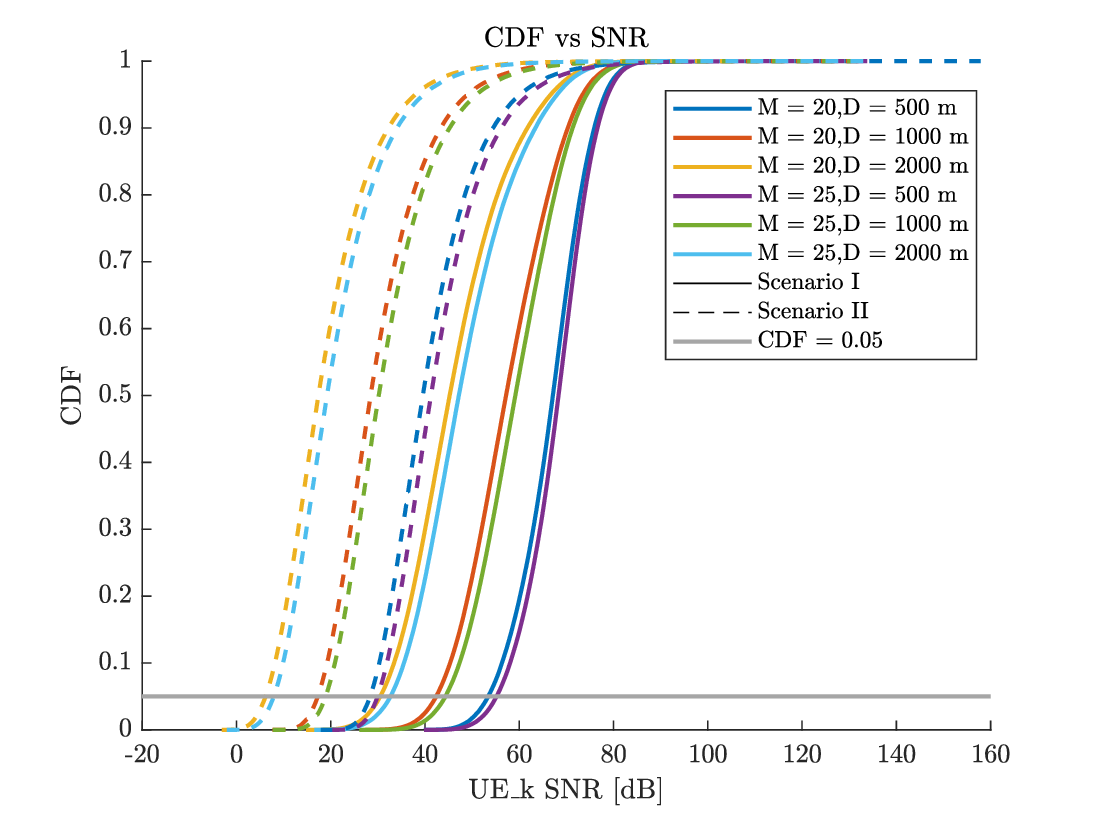}
  \vspace{-20pt} 
  \caption{Active device SNR.}
  \label{fig7}
\end{figure}
\vspace{-15pt}
\protect\subsection{SNR Analysis Results}
\noindent Figure \ref{fig7} illustrates the Cumulative Distribution Function (CDF) of SNR for an active device in a CF-mMIMO network under two scenarios. The SNR is evaluated for devices transmitting at 200 mW across three coverage area sizes specifically 0.5 $\times$ 0.5 km\textsuperscript{2}, 1 $\times$ 1 km\textsuperscript{2}, and 2 $\times$ 2 km\textsuperscript{2}, and with two different numbers of APs, 20 and 25.

 \indent It is clear from Fig.~\ref{fig7} that the SNR achieved in the primary scenario, denoted by solid lines, substantially surpasses that of the secondary scenario from \cite{ganesan2020clustering}, represented by dashed lines. This discrepancy signals a reduced likelihood of outage probability in the primary scenario, which directly translates into more reliable service for a greater number of devices. The superior SNR in the primary scenario can be attributed to a meticulously designed topology of APs and a more realistic environmental model, allowing devices to typically be in closer proximity to an AP, thereby benefiting from stronger channel gains and enhanced SNR.

\indent For simulation purposes, the SNR target at the dominant AP was set to ensure that 95\% of the active devices meet the required SNR threshold to gain network access. This SNR target is depicted by the intersection of the plot lines with the reference line marked $CDF = 0.05$. Based on the graph, the SNR threshold is consistently higher in the $3GPP$ scenario compared to the secondary scenario.

\indent This analysis underscores the advantage of the detailed AP deployment in the primary scenario, emphasizing its potential to enhance connectivity and reduce the risk of signal outage, which is crucial for robust and reliable  performance. Colors differentiate network configurations by area size and AP count, highlighting the impact of these variables on SNR and network performance.
\begin{figure}[h]
  \centering
  \includegraphics[width=\columnwidth]{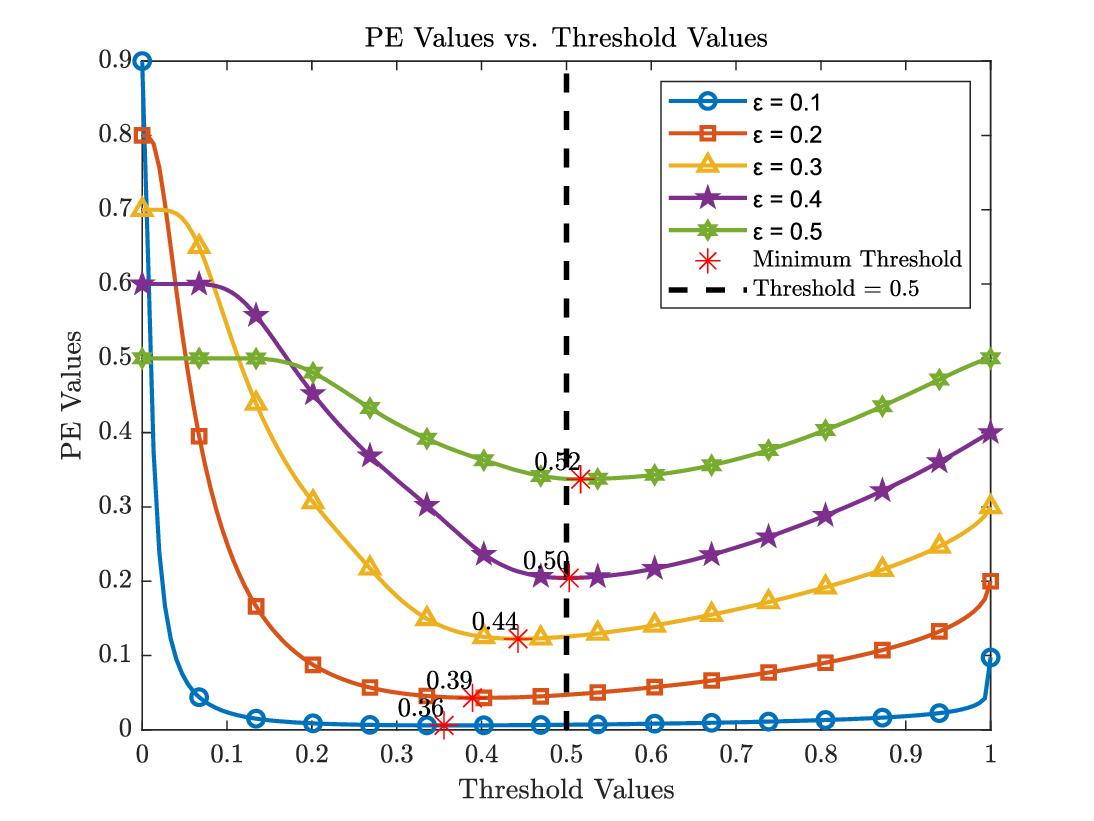}
  \vspace{-20pt} 
  \caption{Probability of error versus threshold values.}
  \label{fig8}
\end{figure}
\protect\subsection{Optimization of the Hard Decision Threshold}
\noindent Figure  \ref{fig8} illustrates the relationship between the probability of error \(P_E (\tau, \epsilon)\) and the decision threshold \(\tau\) for various values of the probability of activation \(\epsilon\) in the context of the DL algorithms performance. Each curve represents the trade-off between FA and MD probabilities, weighted by \(\epsilon\) and \(1-\epsilon\), respectively, and indicates an optimal \(\tau\) minimizing errors for a given \(\epsilon\) that should be used for the most accurate hard decisions made by the DMLP algorithm.

\indent However, given the weighting variables in the error probability equation (\ref{eq:Probability_Error}), it makes logical sense that as $\epsilon$ grows, the ideal threshold $\tau$ increases as well (figure \ref{fig8}). It is interesting to note that error rates reflect very little variation throughout a broad range of threshold values $\tau$, which is centered around $0.5$. This suggests that variations in $\tau$ have minimal impact on the error rate within this range, which may be seen as optimal and remained near optimal for all $\sigma$ values, leading to a simplified hard decision threshold selection methodology.

\begin{figure}[h]
  \centering
  \includegraphics[width=\columnwidth]{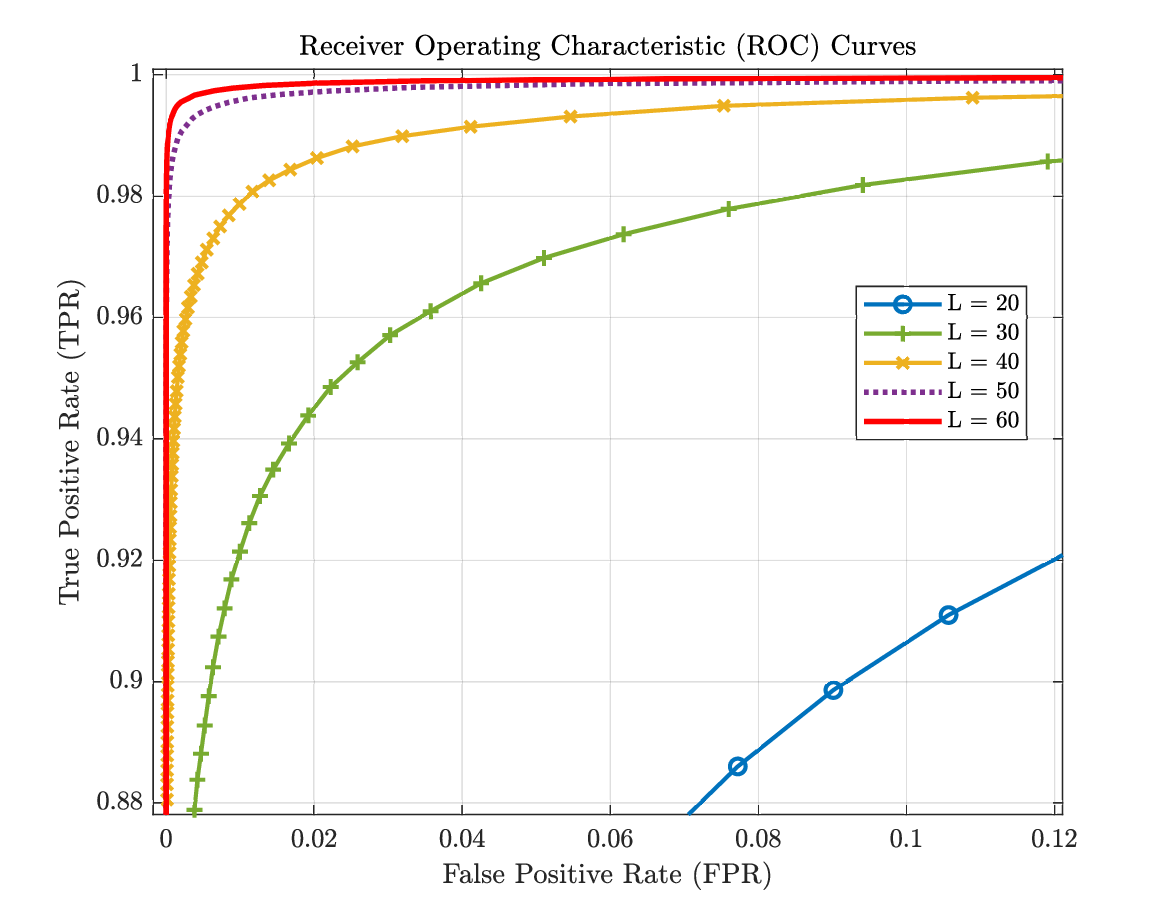}
  \vspace{-20pt} 
  \caption{ROC curve analysis for different pilot sequence lengths \( L \) in DMLP.}
  \label{fig9}
\end{figure}

\indent For instance, even though the precise minimizing thresholds vary for every $\epsilon$, figure \ref{fig8} implies that a standard $\tau$ of $0.5$ might be a useful option over a range of $\epsilon$ without significantly compromising performance. As long as the final performance stays within allowable margins from the absolute minimum error rate, this observation could simplify the parameter selection process in practical situations where selecting a universal threshold may be beneficial for reliability or simplicity without the need for significant calibration.

\begin{figure}[h]
  \centering
  \includegraphics[width=\columnwidth]{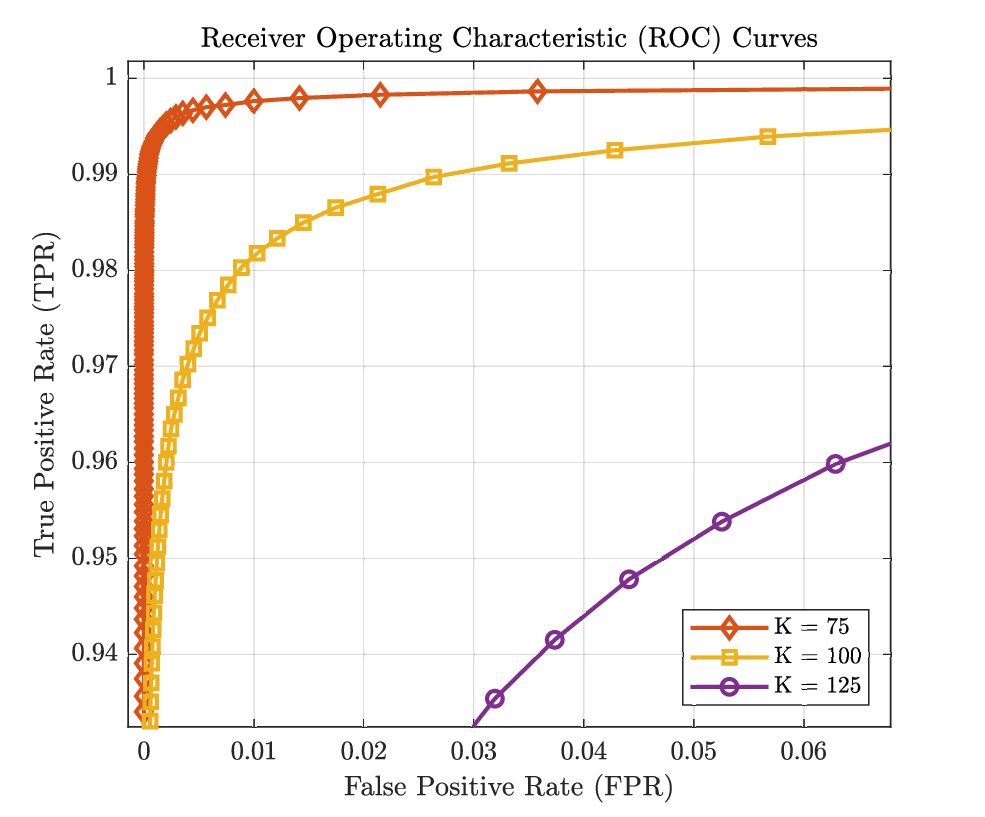}
  \vspace{-10pt} 
  \caption{ROC curve analysis for different user numbers \(K \) in DMLP.}
  \label{fig10}
\end{figure}
\vspace{-10pt}
\protect\subsection{Algorithm Performance with Variable System Parameters}
\noindent In this section, we will be focusing on evaluating our proposed algorithm's performance characteristics as a function of system parameters. Specifically, we will examine the length $L$ of the pilot sequence and the total number of users $K$. These parameters are essential for optimizing the system and provide valuable insights into the adaptability and scalability of the algorithm in real-world deployment scenarios.

\indent The ROC curves for various pilot sequence lengths $L$ are shown in figure \ref{fig9}, and they show that performance improves as the length of the pilot sequence increases from 20 to 60. This non-surprising improvement is attributed to the decrease in correlation between the non-orthogonal pilot sequences given to a set number of users, which makes sense given that the DMLP algorithm improves user discrimination. Longer pilot sequences take up more coherence block space but improve detection performance by lowering sequence correlation. As fewer symbols are available for uplink and downlink data transmission, this consumption is a trade-off that could have an impact on the total data rate and increase the cost of data transmission bandwidth.

\indent Figure \ref{fig10} looks at the DMLP performance as the number of users $K$ increases. As illustrated, performance tends to decrease with increasing user density. This is to be expected since raising $K$ will naturally enhance the correlation between pilot sequences because more users will share a fixed pool of resources, keeping the length of the pilot sequence constant. Referring to the neural network architecture (figure \ref{fig3}) in which the number of users is represented by the number of neurons in the last layer. A static input size and a high user count make it more difficult for the network to discriminate between user signals, which lowers discrimination performance.

\begin{figure}[h]
  \centering
  \includegraphics[width=\columnwidth]{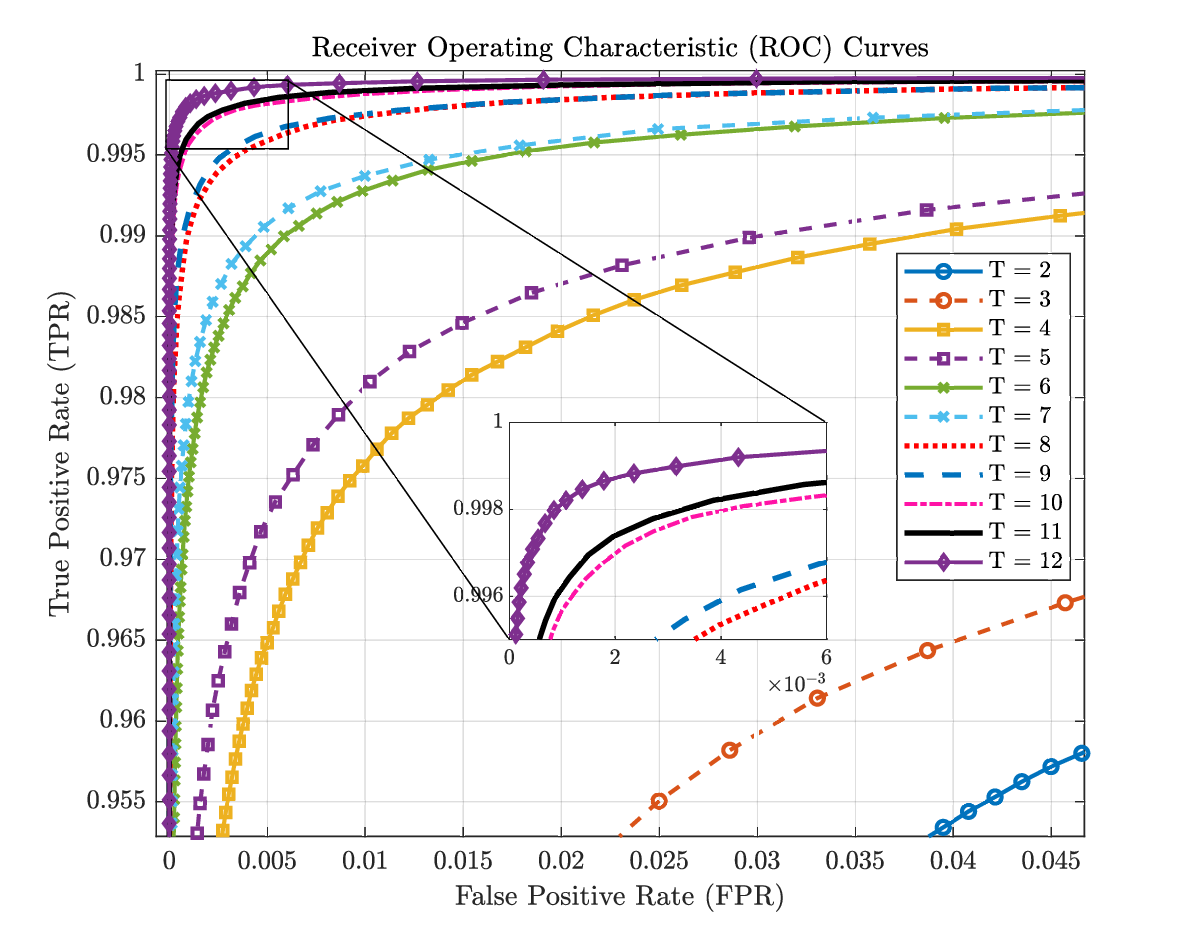}
  \vspace{-15pt} 
  \caption{ROC curve analysis for varying cluster size $T$ in DMLP activity detection.}
  \label{fig11}
\end{figure}

\indent Above all, this scenario illustrates the under-sampling ratio (\(\frac{L}{K}\)), or the trade-off between the number of users $K$ and the length of the pilot sequence $L$. The performance of the algorithm can be optimized by establishing a balance between these variables, highlighting the significance of a comprehensive system design that takes into account the user density and available sequence lengths for a best system operation.

\protect\subsection{Cluster Size Impact on Activity Detection Performance} 
\noindent Figure \ref{fig11} showcases the impact of cluster size ($T$) on the performance of the DMLP algorithm. The ROC curves demonstrate that as the number of AP in a cluster increases, detection performance also improves. 
This enhancement in performance is achieved without the added complexity of solving high-degree polynomial equations, as required by the mathematical algorithm in \cite{ganesan2020clustering}. The traditional approach integrates clustering directly into the algorithm. In contrast, the proposed method separates clustering from the DMLP algorithm. Clustering occurs post-detection, allowing for straightforward majority decisions at the CPU efficiently using simple logical operations. This separation simplifies the overall procedure while preserving the integrity and accuracy of the detection outcome.

\indent According to the analysis (figure \ref{fig11}), there is a slight difference in accuracy based on whether the cluster size is odd or even. When the number of  AP is even, there is a clear majority decision, which reduces the possibility of errors in decision-making. As a result, the accuracy is higher, and the chances of falls due to decision errors are minimized.

\indent When there are an odd number of AP, the majority rule still applies. However, the absence of ties between the AP may allow for a more refined and perceptive detection system, which could account for the slight differences in the ROC curve.

\vspace{-10pt}
\begin{figure}[!t]
  \centering
  \includegraphics[width=\columnwidth]{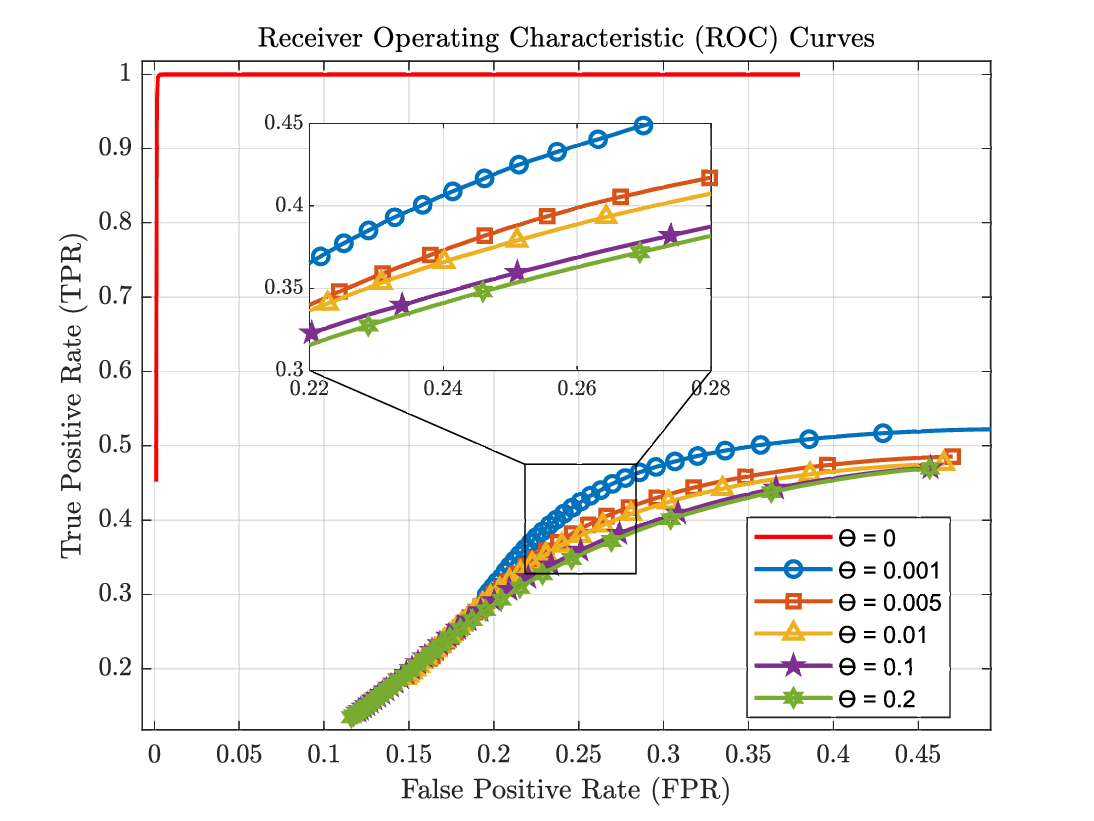}
  \vspace{-15pt} 
  \caption{Mathematical model performance decline from large-scale fading perturbation}
  \label{fig12}
\end{figure}

\protect\subsection{Analysis of Model Resilience Under Imperfect Channel Estimation.}
\noindent In this investigation, we concentrate on the impact of channel estimation errors, which are one of the primary perturbations that can drastically affect the overall performance of any algorithm. This research supports our investigation into another type of disturbance, quantization effects, which will be discussed in a later section. Our focus here is on the consequences of poor channel estimate, using a revised perturbation model derived from reference \cite{Wag2012Pre} to closely imitate real-world conditions:
\begin{equation}
\label{eq:Perturbation}
X_{\text{Pert}} = X  \sqrt{1 - \theta^2} + \theta  n,
\end{equation}

\indent where \(X_{\text{Pert}}\) represents the perturbed version of the input \(X\), which in this context refers to the large-scale fading \(\beta\). The noise term \(n\) is modeled to follow a normal distribution, mirroring the mean and the standard deviation of the original input signal \(\beta\) . The modulation factor \(\theta\), ranging between 0 and 1, dictates the level of original input preservation versus the introduced noise.

\indent This equation models the impact of poor channel estimate on the large-scale fading parameter \(\beta\), which affects the accuracy of the mathematical model. Although we recognize that the equation has little practical relevance when applied to the received signal Y in both the DMLP and the model presented in \cite{ganesan2020clustering}, we continue with this method for completeness. The modest oscillations in $\mathbf{Y}$, which are based on real data, indicate that applying the equation to $\mathbf{Y}$ may not give significant new insights. Nonetheless, this methodological step is critical for determining how each model reacts to changes in its inputs, stressing the equation's relevance in replicating channel estimate errors rather than predicting specific changes in $\mathbf{Y}$.

\indent For the DMLP, perturbing the input $\mathbf{Y}$ does not significantly alter its performance, underscoring the ability of the algorithm to maintain consistency despite input signal fluctuations. This resilience is mirrored in the mathematical model handling of $\mathbf{Y}$ perturbations, where it exhibits minor performance degradation, maintaining a commendable level of accuracy and indicating a degree of robustness, albeit not as pronounced as the DMLP.

\indent The critical evaluation unfolds with figure \ref{fig12}, which illustrates the consequences of disrupting \(\beta\), the second input of the mathematical model. An increase in \(\theta\) leads to a marked performance decline, evidenced by diverging ROC curves. This degradation underscores the dependency of the model on precise large-scale fading information, contrasting sharply with the DMLP's operational efficiency even in the absence of accurate \(\beta\) values. This advantage is paramount in scenarios where accurate channel estimation proves challenging.

\indent The discussion elucidates that for the DMLP, the primary input of concern is the received signal $\mathbf{Y}$, with the algorithm showing remarkable stability across various levels of input perturbation. This is opposed to the mathematical model, which, as figure \ref{fig12} reveals, is significantly affected by inaccuracies in channel estimation, particularly in the estimation of large-scale fading \(\beta\). The reliance on \(\beta\) for the mathematical model highlights its vulnerability to performance degradation under poor channel estimation, a challenge not faced by the DMLP, underscoring the latter superior adaptability and robustness in real-world applications where precise input parameters may not always be ascertainable.

\begin{figure}[h]
  \centering
  \includegraphics[width=\columnwidth]{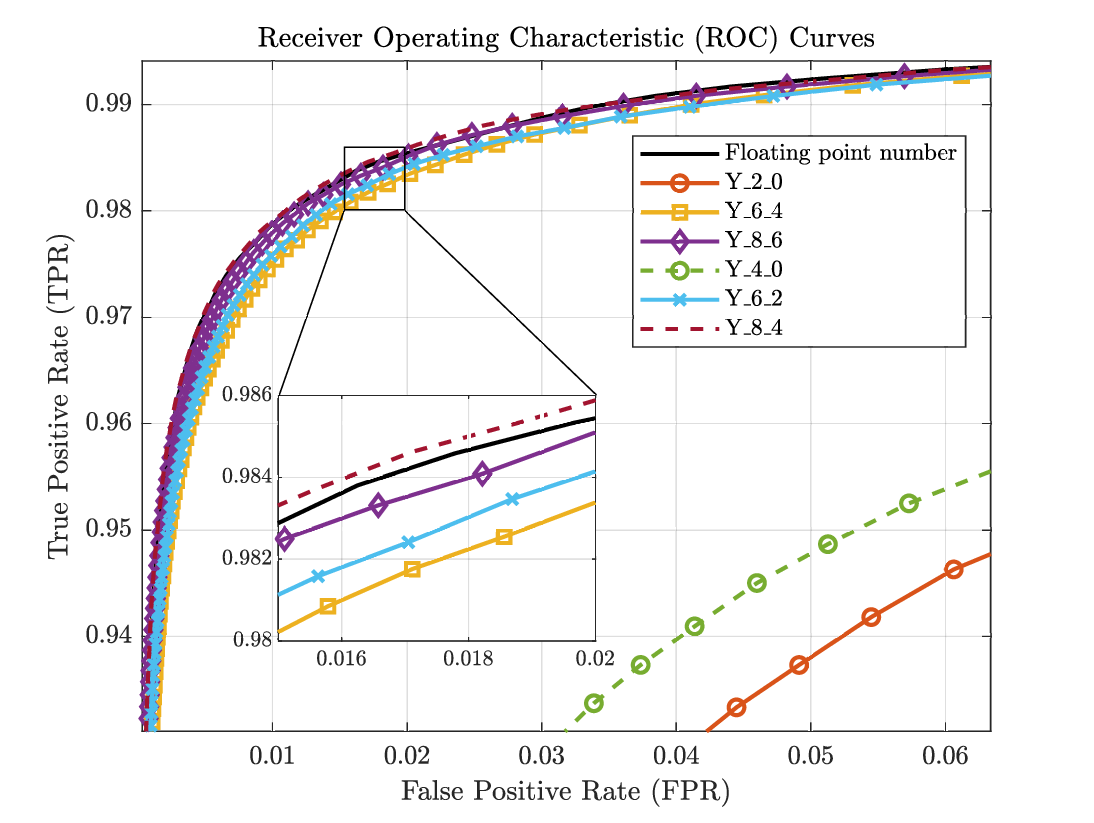}
  \vspace{-15pt} 
  \caption{DMLP performance variation with Fixed-Point Word Lengths for received signal \( Y \), where \( Y\_a\_b \) represents configurations with \( a \) word length and \( b \) fractional bits.}
  \label{fig13}
\end{figure}  

\protect\subsection{Assessing Fixed-Point Quantization Effects on Algorithmic Performance} 
\noindent The conversion from floating-point to fixed-point representation is a critical operation in the concept of  digital signal processing. This transition involves assigning a predetermined number of bits to the integer portion of a signal while the remaining bits represent the fractional part. Such quantization is essential for digital processing and is particularly pertinent in the context of analog-to-digital conversion (ADC). Within this framework, we scrutinize the implications of fixed-point conversion on the performance of two distinct algorithmic approaches: the DMLP and a traditional mathematical model. The DMLP inherently processes the received signal (denoted as $\mathbf{Y}$), whereas the mathematical model contemplates an additional input, the large-scale fading parameter ($\beta$). 

\indent An obvious pattern shows from analyzing the DMLP performance, as shown in figure \ref{fig13}: the ROC curve increases with respect to the word length, which includes both the integer and fractional parts.This suggests that performance will improve as quantization resolution rises. Further investigation indicates that the technique performs better when the integer part is increased while the fractional part bit count remains unchanged. This gain is based on the greater precision for representing larger numbers, which is essential to the correctness of the algorithm and is not random. Similarly, the performance of the DMLP improves when we fix the integer part and increase the precision by adding additional bits for the fractional part. These observations underscore the DMLP sensitivity to both the resolution of the integer and the precision of the fractional parts.

\begin{figure}[h]
  \centering
  \includegraphics[width=\columnwidth]{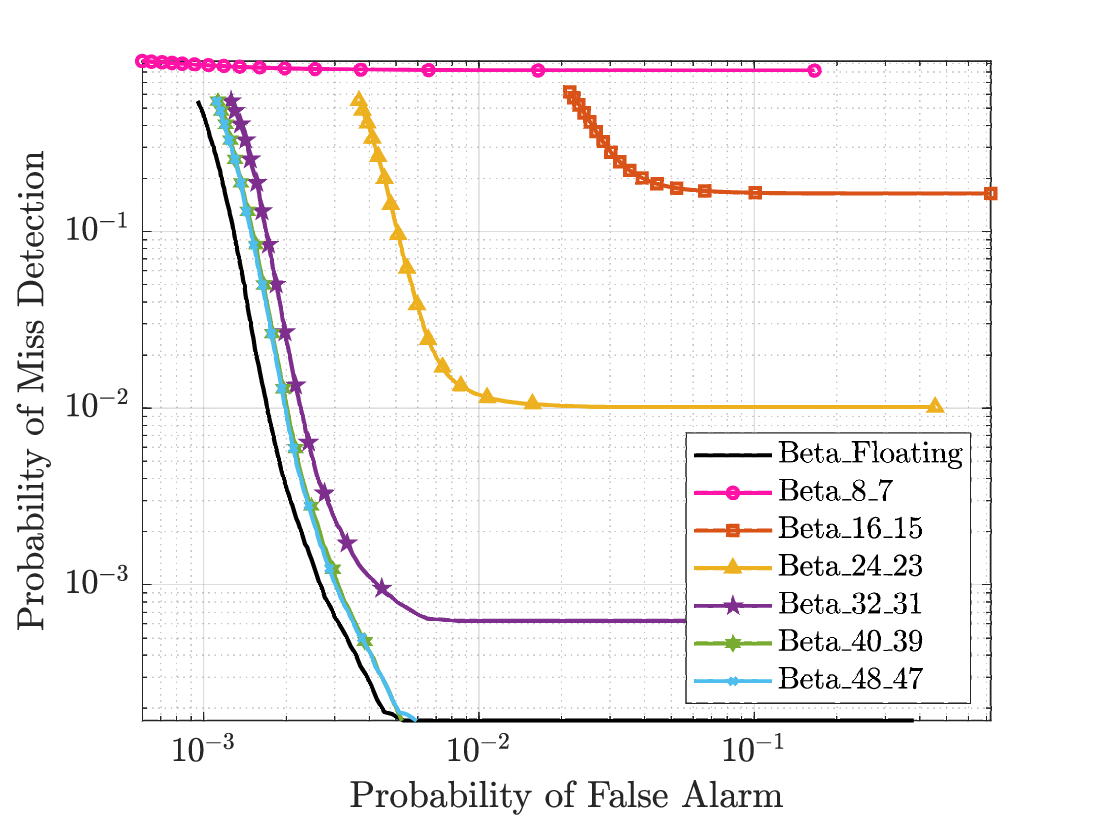}
  \vspace{-15pt} 
  \caption{Impact of Fixed-Point large-scale fading $\beta$ on mathematical model Performance, where \( Beta\_a\_b \) represents configurations with \( a \) word length and \( b \) fractional bits.}
  \label{fig14}
\end{figure}

\indent The mathematical model performance, utilizing a fixed-point representation for only the large-scale fading input \( \beta \), is thoroughly analyzed. The extremely small values of \( \beta \), on the order of \( 10^{-13} \), require a high level of precision for accurate representation. Figure \ref{fig14} shows that the model's performance improves as the precision of the input, or the number of bits in the decimal part, increases. A noticeable decrease in accuracy with a reduction in the number of decimal bits highlights the importance of precise \( \beta \) representation in fixed-point format.

\indent Additionally, it is noted that the curves closest to the model performance with floating-point representation without any conversion are achieved at a precision level of 48 bits (6 Bytes). This precision level, while enabling closer performance to floating-point accuracy, is significantly high and comes with several disadvantages. These include increased memory demands, higher computational resources, and greater power consumption. Such requirements can lead to a more complex system design, potentially affecting the system overall efficiency and practicality, especially in environments where resources are limited. Selecting the appropriate level of precision is vital for balancing performance improvements against these challenges.

\indent In summary, the DMLP experiences uniform benefits from heightened precision in both integer and fractional segments. Conversely, the mathematical model displays a notable preference contingent upon the input undergoing fixed-point conversion, signaling a nuanced and varied reaction to fixed-point conversion between the two algorithmic paradigms.

\section{Conclusion}
\label{sec:Conclusion}
\noindent This paper presented a thorough examination of grant-free random access protocols within cell-free massive MIMO networks, where a DL algorithm named DMLP was introduced for efficient sparse support recovery and active device detection. With an emphasis on low computational complexity and rapid execution times, following the initial training phase, the study showcased the superior performance of the algorithm through various simulations. These included perturbation analysis and fixed-point conversion. 
The results showed a significant improvement over existing approaches,highlighting the stability of cell-free massive MIMO networks in improving coverage for mMTC applications. This work provided a possible solution to the issues of huge connectivity in future wireless networks.

\section*{Acknowledgements}
\noindent This work received funding from the French National Research Agency (ANR-22-CE25-0015) within the frame of the project POSEIDON.

\bibliography{References}

\begin{thebibliography}{10}
\providecommand{\url}[1]{#1}
\csname url@samestyle\endcsname
\providecommand{\newblock}{\relax}
\providecommand{\bibinfo}[2]{#2}
\providecommand{\BIBentrySTDinterwordspacing}{\spaceskip=0pt\relax}
\providecommand{\BIBentryALTinterwordstretchfactor}{4}
\providecommand{\BIBentryALTinterwordspacing}{\spaceskip=\fontdimen2\font plus
\BIBentryALTinterwordstretchfactor\fontdimen3\font minus \fontdimen4\font\relax}
\providecommand{\BIBforeignlanguage}[2]{{%
\expandafter\ifx\csname l@#1\endcsname\relax
\typeout{** WARNING: IEEEtran.bst: No hyphenation pattern has been}%
\typeout{** loaded for the language `#1'. Using the pattern for}%
\typeout{** the default language instead.}%
\else
\language=\csname l@#1\endcsname
\fi
#2}}
\providecommand{\BIBdecl}{\relax}
\BIBdecl

\bibitem{Bockelmann2016}
C.~Bockelmann, N.~Pratas, H.~Nikopour, K.~Au, T.~Svensson, C.~Stefanovic, P.~Popovski, and A.~Dekorsy, ``Massive machine-type communications in 5g: physical and mac-layer solutions,'' \emph{IEEE Communications Magazine}, vol.~54, no.~9, pp. 59--65, 2016.

\bibitem{Bana2019}
A.-S. Bana, E.~de~Carvalho, B.~Soret, T.~Abrão, J.~C. Marinello, E.~G. Larsson, and P.~Popovski, ``Massive mimo for internet of things (iot) connectivity,'' 2019.

\bibitem{Marzetta2010}
T.~L. Marzetta, ``Noncooperative cellular wireless with unlimited numbers of base station antennas,'' \emph{IEEE Transactions on Wireless Communications}, vol.~9, no.~11, pp. 3590--3600, 2010.

\bibitem{Pratas2012}
N.~K. Pratas, H.~Thomsen, C.~Stefanović, and P.~Popovski, ``Code-expanded random access for machine-type communications,'' in \emph{2012 IEEE Globecom Workshops}, 2012, pp. 1681--1686.

\bibitem{Sorensen2014}
J.~H. Sørensen, E.~de~Carvalho, and P.~Popovski, ``Massive mimo for crowd scenarios: A solution based on random access,'' in \emph{2014 IEEE Globecom Workshops (GC Wkshps)}, 2014, pp. 352--357.

\bibitem{Bjornson2016}
E.~Björnson, E.~de~Carvalho, E.~G. Larsson, and P.~Popovski, ``Random access protocol for massive mimo: Strongest-user collision resolution (sucr),'' in \emph{2016 IEEE International Conference on Communications (ICC)}, 2016, pp. 1--6.

\bibitem{Ding2020}
J.~Ding, D.~Qu, and J.~Choi, ``Analysis of non-orthogonal sequences for grant-free ra with massive mimo,'' \emph{IEEE Transactions on Communications}, vol.~68, no.~1, pp. 150--160, 2020.

\bibitem{Shahab2020}
M.~B. Shahab, R.~Abbas, M.~Shirvanimoghaddam, and S.~J. Johnson, ``Grant-free non-orthogonal multiple access for iot: A survey,'' \emph{IEEE Communications Surveys \& Tutorials}, vol.~22, no.~3, pp. 1805--1838, 2020.

\bibitem{Chen2021}
X.~Chen, D.~W.~K. Ng, W.~Yu, E.~G. Larsson, N.~Al-Dhahir, and R.~Schober, ``Massive access for 5g and beyond,'' \emph{IEEE Journal on Selected Areas in Communications}, vol.~39, no.~3, pp. 615--637, 2021.

\bibitem{bockelmann2013compressive}
\BIBentryALTinterwordspacing
C.~Bockelmann, H.~F. Schepker, and A.~Dekorsy, ``Compressive sensing based multi-user detection for machine-to-machine communication,'' \emph{Transactions on Emerging Telecommunications Technologies}, vol.~24, no.~4, pp. 389--400, 2013. [Online]. Available: \url{https://onlinelibrary.wiley.com/doi/abs/10.1002/ett.2633}
\BIBentrySTDinterwordspacing

\bibitem{monsees2015compressive}
F.~Monsees, M.~Woltering, C.~Bockelmann, and A.~Dekorsy, ``Compressive sensing multi-user detection for multicarrier systems in sporadic machine type communication,'' in \emph{2015 IEEE 81st Vehicular Technology Conference (VTC Spring)}, 2015, pp. 1--5.

\bibitem{gao2015compressive}
Z.~Gao, L.~Dai, Z.~Wang, S.~Chen, and L.~Hanzo, ``Compressive-sensing-based multiuser detector for the large-scale sm-mimo uplink,'' \emph{IEEE Transactions on Vehicular Technology}, vol.~65, no.~10, pp. 8725--8730, 2016.

\bibitem{du2017efficient}
Y.~Du, B.~Dong, Z.~Chen, X.~Wang, Z.~Liu, P.~Gao, and S.~Li, ``Efficient multi-user detection for uplink grant-free noma: Prior-information aided adaptive compressive sensing perspective,'' \emph{IEEE Journal on Selected Areas in Communications}, vol.~35, no.~12, pp. 2812--2828, 2017.

\bibitem{liu2018massive}
L.~Liu and W.~Yu, ``Massive connectivity with massive mimo—part i: Device activity detection and channel estimation,'' \emph{IEEE Transactions on Signal Processing}, vol.~66, no.~11, pp. 2933--2946, 2018.

\bibitem{senel2018grant}
K.~Senel and E.~G. Larsson, ``Grant-free massive mtc-enabled massive mimo: A compressive sensing approach,'' \emph{IEEE Transactions on Communications}, vol.~66, no.~12, pp. 6164--6175, 2018.

\bibitem{senel2017device}
------, ``Device activity and embedded information bit detection using amp in massive mimo,'' in \emph{2017 IEEE Globecom Workshops (GC Wkshps)}, 2017, pp. 1--6.

\bibitem{donoho2009message}
\BIBentryALTinterwordspacing
D.~L. Donoho, A.~Maleki, and A.~Montanari, ``Message-passing algorithms for compressed sensing,'' \emph{Proceedings of the National Academy of Sciences}, vol. 106, no.~45, p. 18914–18919, Nov. 2009. [Online]. Available: \url{http://dx.doi.org/10.1073/pnas.0909892106}
\BIBentrySTDinterwordspacing

\bibitem{HaghiJungCaire2018}
S.~Haghighatshoar, P.~Jung, and G.~Caire, ``Improved scaling law for activity detection in massive mimo systems,'' in \emph{2018 IEEE International Symposium on Information Theory (ISIT)}, 2018, pp. 381--385.

\bibitem{chen2019covariance}
Z.~Chen, F.~Sohrabi, Y.-F. Liu, and W.~Yu, ``Covariance based joint activity and data detection for massive random access with massive mimo,'' in \emph{ICC 2019 - 2019 IEEE International Conference on Communications (ICC)}, 2019, pp. 1--6.

\bibitem{ganesan2020clustering}
U.~K. Ganesan, E.~Björnson, and E.~G. Larsson, ``Clustering-based activity detection algorithms for grant-free random access in cell-free massive mimo,'' \emph{IEEE Transactions on Communications}, vol.~69, no.~11, pp. 7520--7530, 2021.

\bibitem{Ganesan2020Algorithm}
U.~K. Ganesan, E.~Bjornson, and E.~G. Larsson, ``An algorithm for grant-free random access in cell-free massive mimo,'' in \emph{2020 IEEE 21st International Workshop on Signal Processing Advances in Wireless Communications (SPAWC)}, 2020, pp. 1--5.

\bibitem{deSouza2023DeepLearning}
J.~H.~I. de~Souza and T.~Abrão, ``Deep learning-based activity detection for grant-free random access,'' \emph{IEEE Systems Journal}, vol.~17, no.~1, pp. 940--951, 2023.

\bibitem{bai2018deep}
Y.~Bai, B.~Ai, and W.~Chen, ``Deep learning based fast multiuser detection for massive machine-type communication,'' in \emph{2019 IEEE 90th Vehicular Technology Conference (VTC2019-Fall)}, 2019, pp. 1--5.

\bibitem{3gppTR38901}
\BIBentryALTinterwordspacing
{3rd Generation Partnership Project}, ``Study on channel model for frequencies from 0.5 to 100 ghz,'' 3GPP TR 38.901 V17.0.0 (2022-03), 3GPP, Valbonne, France, Technical Report TR 38.901, March 2022, release 17. [Online]. Available: \url{http://www.3gpp.org}
\BIBentrySTDinterwordspacing

\bibitem{Nam2014LargeScale}
\BIBentryALTinterwordspacing
J.~Nam, J.~Kim, E.~Loza~Mencía, I.~Gurevych, and J.~Fürnkranz, \emph{Large-Scale Multi-label Text Classification — Revisiting Neural Networks}.\hskip 1em plus 0.5em minus 0.4em\relax Springer Berlin Heidelberg, 2014, p. 437–452. [Online]. Available: \url{http://dx.doi.org/10.1007/978-3-662-44851-9_28}
\BIBentrySTDinterwordspacing

\bibitem{Wag2012Pre}
S.~Wagner, R.~Couillet, M.~Debbah, and D.~T.~M. Slock, ``Large system analysis of linear precoding in correlated miso broadcast channels under limited feedback,'' \emph{IEEE Transactions on Information Theory}, vol.~58, no.~7, pp. 4509--4537, 2012.

\end{thebibliography}
\bibliographystyle{IEEEtran}

\end{document}